\documentclass[sigconf]{acmart}

\usepackage{amsmath,amsfonts,bm}

\def\eqref#1{equation~\ref{#1}}

\def\1{\bm{1}}

\DeclareMathAlphabet{\mathsfit}{\encodingdefault}{\sfdefault}{m}{sl}
\SetMathAlphabet{\mathsfit}{bold}{\encodingdefault}{\sfdefault}{bx}{n}

\usepackage{bbm}
\usepackage{algorithm}
\usepackage{algorithmic}

\usepackage{booktabs}
\usepackage{multirow}
\usepackage{graphicx}
\usepackage{xcolor}
\usepackage{colortbl}

\AtBeginDocument{%
  }

\setcopyright{acmlicensed}
\copyrightyear{2018}
\acmYear{2018}
\acmDOI{XXXXXXX.XXXXXXX}
\acmConference[Conference acronym 'XX]{Make sure to enter the correct
  conference title from your rights confirmation email}{June 03--05,
  2018}{Woodstock, NY}
\acmISBN{978-1-4503-XXXX-X/2018/06}

\begin{document}

\title{Generative Recall, Dense Reranking: Learning Multi-View Semantic IDs for Efficient Text-to-Video Retrieval}

\author{Zecheng Zhao}
\affiliation{%
  \institution{The University of Queensland}
  \city{Brisbane}
  \country{Australia}}
\email{uqzzha35@uq.edu.au}

\author{Zhi Chen}
\affiliation{%
  \institution{The University of Southern Queensland}
  \city{Toowoomba}
  \country{Australia}}
\email{zhi.chen@unisq.edu.au}

\author{Zi Huang}
\affiliation{%
  \institution{The University of Queensland}
  \city{Brisbane}
  \country{Australia}}
\email{huang@itee.uq.edu.au}

\author{Shazia Sadiq}
\affiliation{%
  \institution{The University of Queensland}
  \city{Brisbane}
  \country{Australia}}
\email{shazia@eecs.uq.edu.au}

\author{Tong Chen}
\authornote{Corresponding author.}
\affiliation{%
  \institution{The University of Queensland}
  \city{Brisbane}
  \country{Australia}}
\email{tong.chen@uq.edu.au}

\begin{abstract}
Text-to-Video Retrieval (TVR) is an essential service in video platforms. In TVR, dense retrieval with dual-modality encoders leads in accuracy, but its computation and storage scale poorly with the corpus size. Thus, real-time and large-scale applications commonly adopt two-stage retrieval, where a fast recall model gathers a small pool of candidates, which are reranked by an advanced dense retriever. Due to hugely reduced video candidates, the reranking model can use any off-the-shelf dense retriever without hurting efficiency, meaning that the recall model ultimately bounds the performance of two-stage TVR. Recently, the emerging paradigm of generative retrieval (GR) replaces dense video embeddings with discrete semantic IDs, and retrieves by decoding the text queries into ID tokens. GR offers near-constant inference and storage complexity, and its semantic IDs capture high-level video features via quantization, making it ideal for quickly eliminating irrelevant candidates during recall. However, when serving as a recall model in two-stage TVR, GR suffers from (i) semantic ambiguity, where each video satisfies diverse queries but is forced into one semantic ID; and (ii) cross-modal misalignment, as semantic IDs are solely derived from videos' visual features without any supervision from text queries. In this paper, we propose the Generative Recall and Dense Reranking (GRDR) framework, in which we design a novel GR method to uplift the quality of recalled candidates. GRDR assigns multiple semantic IDs to each video using a query-guided multi-view tokenizer to expose diverse semantic access paths, and jointly trains the tokenizer and generative retriever via a shared codebook to cast semantic IDs as the semantic bridge between texts and videos. At inference, trie-constrained decoding generates a compact candidate set that is then reranked by a dense model, ensuring fine-grained matching. Experiments on four TVR benchmarks show that GRDR is on par with strong dense retrievers in terms of accuracy, while reducing index storage by over an order of magnitude and accelerating by up to 300$\times$ in full-corpus retrieval. Our code is available at: 
\url{https://github.com/JasonCodeMaker/GRDR}
\end{abstract}




\maketitle

\definecolor{online}{HTML}{FF6F61} 
\definecolor{offline}{HTML}{5B5B5B} 

\section{Introduction}


Text-to-Video Retrieval (TVR) \cite{BLIM, TVR_CLIP4clip, GT_TVR_T2VIndexer} aims to locate videos relevant to natural language queries, particularly useful in video platforms that host content at unprecedented scale with billions of videos. At this scale, users demand high retrieval accuracy but still expect a sub-second response time. For service providers, storage and computational costs must remain tractable as corpus size grows. Altogether, these requirements prioritize retrieval paradigms that jointly optimize both effectiveness and efficiency.

\begin{figure}[!t]
    \centering
    \includegraphics[width=1\linewidth]{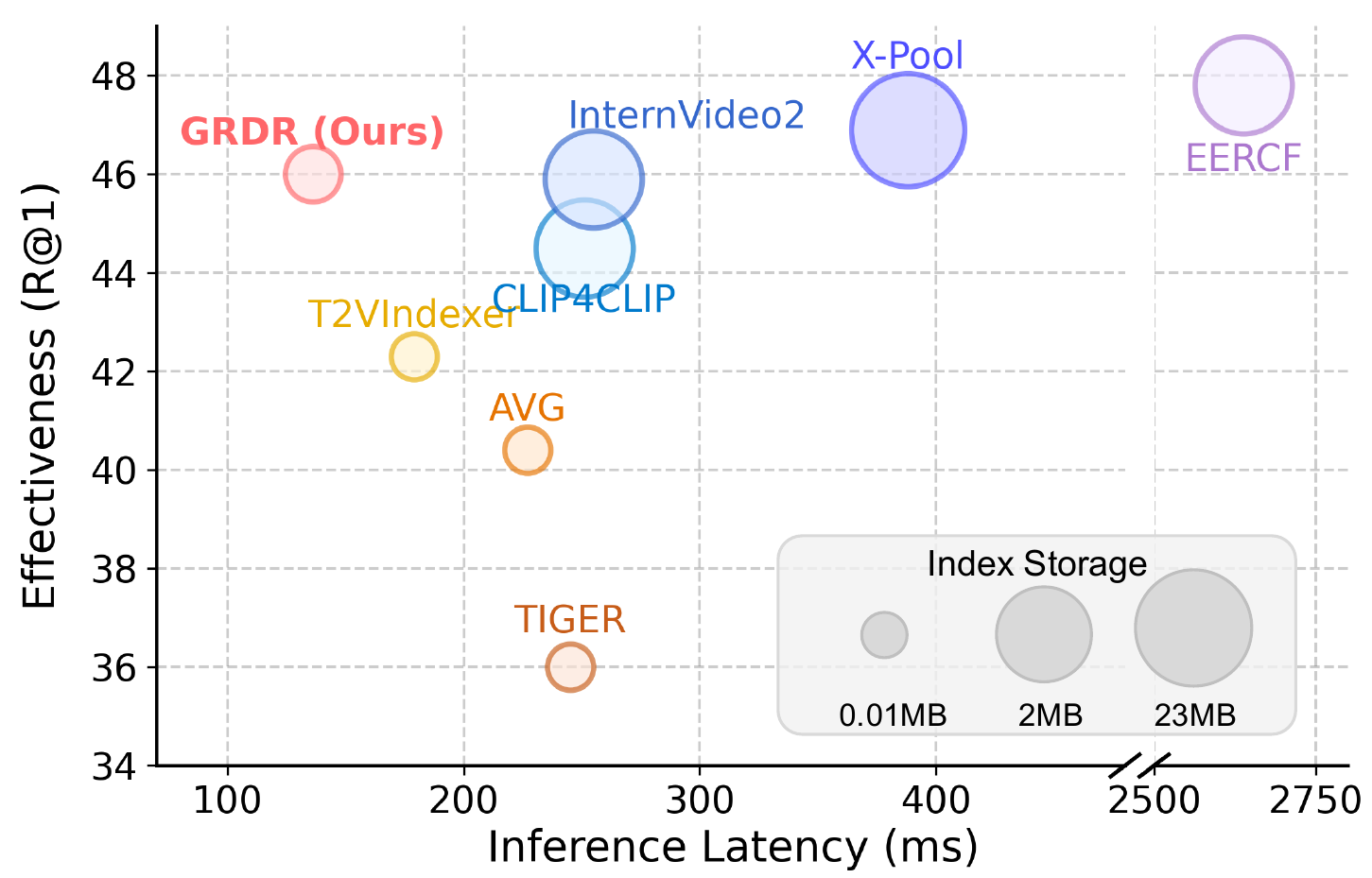}
    \vspace{-2em}
    \caption{Effectiveness vs. Efficiency on MSRVTT-1k test set~\cite{MSRVTT}. Bubble size indicates the index storage required for cached video features. GRDR (Ours) achieves the optimal balance (positioned at the top-left). }
    \label{fig:efficiency_effectiveness}
    \vspace{-1em}
\end{figure}

\begin{figure*}[!t]
    \centering
    \includegraphics[width=1\linewidth]{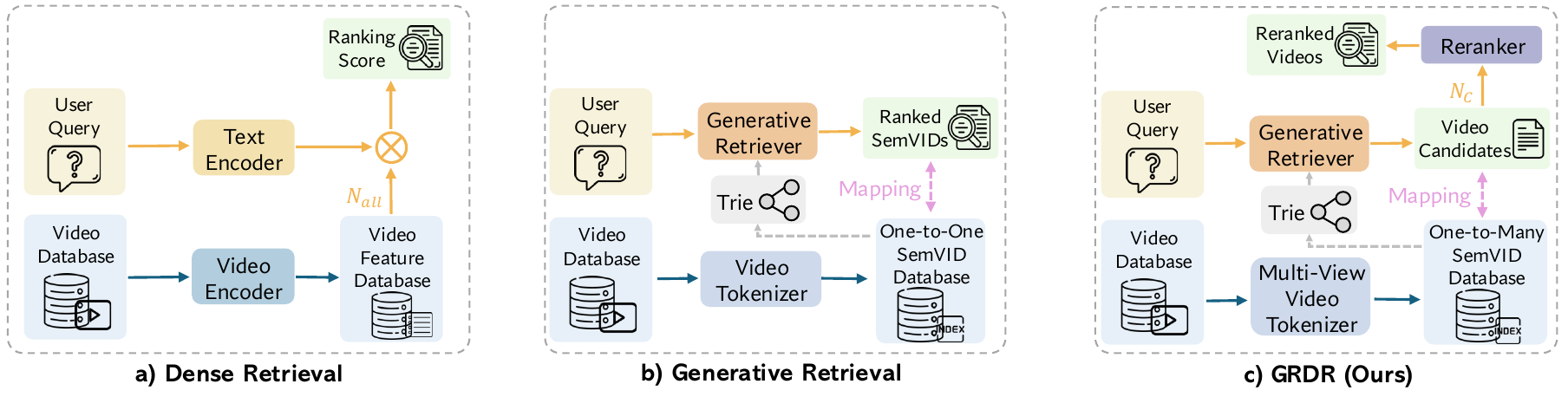}
    \vspace{-15pt}
    \caption{Overview of Text-to-Video Retrieval Paradigms. (a) Dense Retrieval: Decouples modalities into dual-encoders, relying on exhaustive similarity search over high-dimensional embeddings. (b) Generative Retrieval: Reformulates retrieval as sequence-to-sequence generation, constrained by a rigid one-to-one video-to-identifier mapping. (c) GRDR (Ours): leverage a Recall-then-Rerank paradigm using multi-view (one-to-many) tokenization to capture diverse video semantics.}
    \label{fig:paradigms}
    \vspace{-10pt}
\end{figure*}

Dense retrieval \cite{TVR_CLIP4clip, TVR_CLIPViP, TVR_Xpool, InternVideo2} is the predominant paradigm in current TVR systems by decoupling video and text encoding into two branches. As shown in Figure \ref{fig:paradigms}(a), videos are encoded offline as high-dimensional embeddings. At query time, received text queries are encoded as embeddings, and used to score and rank videos by computing pairwise embedding similarity. 
Although dense retrieval achieves optimal retrieval accuracy, this paradigm faces inherent efficiency limitations as the corpus size grows. First, retrieval cost becomes a substantial throughput bottleneck, where high-dimensional video features are recursively loaded into memory for similarity calculation. Second, storage overhead increases linearly, as high-dimensional video embeddings must be retained for the entire video collection.


These efficiency limitations motivate Generative Retrieval (GR) \cite{GR_DR_NCI, GR_DR_DSI, GR_RecSys_TIGER} for TVR. As shown in Figure~\ref{fig:paradigms}(b), GR reformulates retrieval as a sequence-to-sequence generation problem, with (i) a video tokenizer that quantizes each video into a sequence of discrete tokens, namely the semantic ID, and (ii) a generative retriever that auto-regressively decodes target semantic IDs from a query. By replacing video embeddings with semantic IDs, GR achieves near-constant storage complexity and inference efficiency. However, despite strong compatibility with single-modal tasks like document retrieval \cite{GR_DR, GR_DR_1, GR_DR_DSI, GR_DR_NCI}, GR's effectiveness in TVR is constrained compared with dense retrieval. When rich video information is compressed into few discrete tokens, only high-level semantics are retained (\textit{e.g.,} genre, category, key objects), while finer nuances essential for accurate retrieval (\textit{e.g.,} dynamic motion, temporal transitions, fine-grained object interactions) are lost. Consequently, GR's efficiency comes at the cost of suboptimal TVR accuracy.


Meanwhile, large-scale retrieval typically adopts a two-stage recall-and-rerank paradigm~\cite{RecallRerank1, TVR_EERCF} to balance efficiency and effectiveness, which has become a common practice in industrial systems \cite{RecallRerank1, Recall_Rerank2}. The recall stage narrows the full video corpus down to a small candidate pool with a fast model~\cite{LSH,BM25}, prioritizing efficient filtering with coarse information. The rerank stage further refines the ranking, prioritizing accuracy by extracting fine-grained features. During reranking, as only a fraction of the videos are processed, a sophisticated dense retriever can be conveniently utilized without harming the overall efficiency. As such, the performance of two-stage TVR is essentially capped by the quality of recalled candidates, i.e., the recall model needs to correctly place the target video in the candidate set. As GR focuses on summarizing video content into high-level features, this makes GR a natural fit for candidate recall. The primary goal of the recall stage is eliminating the majority of query-irrelevant videos in an efficient manner, whereas the intricate task of differentiating highly alike candidates can be offloaded to the dense reranking model.

In this regard, to achieve efficient yet performant TVR, we explore the technical pathways to integrate GR into the recall process of two-stage TVR. However, existing GR methods \cite{GR_RecSys_TIGER, GR_IR_AVG, GT_TVR_T2VIndexer} are limited by two fundamental challenges, rendering them a suboptimal choice as a recall model:

\noindent \textbf{(i) Semantic Ambiguity.} Videos are inherently polysemous. A video consists of multiple events and temporal dynamics described by diverse text queries. For example, a cooking video is relevant to queries about the prepared dish, the demonstrated technique, or the used utensils. However, existing GR tokenizers \cite{GR_RecSys_TIGER, GR_IR_AVG, GT_TVR_T2VIndexer} enforce a one-to-one mapping from videos to semantic IDs. Thus, queries with distinct intents may converge to identical semantic IDs. This leads to recall failures when the generated path misses a valid semantic facet. 
A naïve solution is to extend the length of semantic IDs to sequentially encode additional semantic facets, yet it defies the parallel nature of those diverse video facets and makes the auto-regressive decoding prone to error accumulation.
Besides, long semantic IDs also incur high inference latency. 



\noindent \textbf{(ii) Cross-Modal Misalignment.} 
Existing GR pipelines \cite{GR_RecSys_TIGER, GR_IR_AVG, GT_TVR_T2VIndexer} train the tokenizer and the generative retriever separately. Under this design, the tokenizer optimizes semantic IDs for visual coherence, while the generative retriever learns to map text queries to corresponding semantic IDs. As supervision signals from the retrieval task do not flow back to the tokenizer, the semantic IDs are misaligned with the generative retriever's textual embedding space where retrieval is performed. Though methods like AVG~\cite{GR_IR_AVG} attempt to address this by incorporating cross-modal features during tokenization, the optimization of semantic IDs remains disconnected from the retrieval objective, weakening GR's efficacy for candidate recall in TVR.

To address these challenges, we propose a Generative Recall and Dense Reranking framework (GRDR) that learns multi-view semantic IDs for efficient text-to-video retrieval. 
Specifically, first, to mitigate semantic ambiguity, we propose a multi-view video tokenizer that encodes each video into multiple semantic IDs. To ensure each encoder captures distinct semantics, we adopt query-guided encoding where each encoder aligns with a specific query intent. This alignment ensures video representations are optimized directly in the retrieval embedding space, enabling polysemous videos to be retrieved through diverse semantic paths.
Second, to resolve cross-modal misalignment, we introduce a unified co-training scheme that couples the tokenizer and the generative retriever through identical codebook embeddings for code selection. This integration achieves a unified semantic space that supports bidirectional gradient flow and end-to-end optimization. We conduct experiments on four TVR benchmarks. Results show that our method outperforms existing generative retrieval baselines and approaches dense retrieval accuracy, while achieving substantial gains in both storage and inference efficiency.

In summary, our contributions are as follows:
\begin{itemize}
    \item We formalize a practical Generative Recall and Dense Reranking (GRDR) paradigm for TVR, where generative retrieval serves as a high-efficiency candidate generator and dense reranking preserves fine-grained matching.
    \item We propose a multi-view video tokenizer that assigns multiple semantic IDs per video via query-guided contrastive learning to handle semantic ambiguity, and a unified co-training scheme that couples tokenizer and retriever in an end-to-end manner, enabling bidirectional gradient flow and allowing retrieval-aware semantic ID assignment to address cross-modal misalignment.
    \item Experiments on four TVR benchmarks show our method maintains competitive retrieval effectiveness while achieving order-of-magnitude storage reduction and up to $300 \times$ speed-up compared with dense retrieval baselines.
\end{itemize}

\vspace{-1em}
\section{Related Work}

\textbf{Text-to-Video Retrieval.}  Text-to-Video Retrieval (TVR) aims to retrieve videos from a corpus based on their semantic relevance to a text query. Existing methods fall into two paradigms distinguished by their cross-modal interaction strategy. 
Cross-Encoder Retrieval \cite{BLIM, InternVideo2, BLIP} leverages deep cross-modal fusion to achieve high retrieval accuracy. With recent advances in large-scale Vision-Language Models (VLMs) \cite{VLM1, VLM2, VLM3, VLM4, CV, CV1, CLIP3}, models such as InternVideo series \cite{InternVideo, InternVideo2, InternVideo2.5} and BLiM \cite{BLIM} employ cross-attention mechanisms to capture fine-grained interactions. However, this architecture requires a full forward pass for every query-video pair at retrieval time, making it impractical for large-scale retrieval.
The Dense Retrieval paradigm addresses this bottleneck by decoupling video and text encoding into separate branches. Videos can be encoded offline, reducing retrieval to nearest-neighbor search over precomputed embeddings. Building on pretrained vision-language models such as CLIP \cite{CLIP}, subsequent methods \cite{TVR_ProST, TVR_Xpool, TVR_CLIPBert, TVR_CLIP4clip, TVR_MV-Adapter, TVR_Token_Shift, TVR_CLIPViP, TVR_DiffusionRet, TVR_Prompt_Switch, TVR_CTVR, TVR_SynTVR, CLIP2} extend this paradigm through temporal modeling and cross-modal alignment mechanisms. Recently, InternVideo2 \cite{InternVideo2} proposes a CLIP-style dual-encoder variant to further improve dense retrieval baselines through model scaling. Despite these advances, Dense Retrieval faces inherent scalability limitations. Storage complexity grows linearly with corpus size due to high-dimensional video features. Inference complexity also scales linearly, as retrieval requires exhaustive similarity computation across all candidates.
EERCF \cite{TVR_EERCF} attempts to mitigate inference costs through Coarse-to-Fine Retrieval. It proposes a two-stage framework that first filters candidates before fine-grained reranking. However, the recall stage still relies on dense representations, inheriting the same storage overhead. These limitations motivate Generative Retrieval, which achieves compact storage and efficient inference through the generation of discrete semantic IDs. 

\noindent \textbf{Generative Retrieval.} Generative Retrieval (GR) \cite{GR_DR, GR_DR_1, GR_IR_AVG, GR_RecSys, GR_RecSys_ETEGRec, GR_RecSys_TIGER, GR_RecSys1, GR_RecSys2, GR_RecSys3, GT_TVR_T2VIndexer, GR_DR_NCI, GR_DR_DSI} reformulates retrieval as a sequence-to-sequence generation task, where models directly produce target identifiers conditioned on queries. Following its foundational success in document retrieval \cite{GR_DR, GR_DR_1, GR_DR_NCI, GR_DR_DSI, GR_DR4}, the paradigm has recently been adapted to recommendation systems \cite{GR_RecSys, GR_RecSys1, GR_RecSys2, GR_RecSys3, GR_RecSys_TIGER, GR_RecSys_ETEGRec}. TIGER \cite{GR_RecSys_TIGER} first proposes semantic IDs using Residual Quantization (RQ) \cite{RQ} to generate hierarchical discrete codes for items in sequential recommendation. In addition, Actionpiece \cite{GR_RecSys} extends this by tokenizing user behaviors using a variable-length mechanism. ETEGRec \cite{GR_RecSys_ETEGRec} further advanced this direction by jointly optimizing item tokenization and recommendation objectives, eliminating the disjoint multi-stage pipeline.
Building upon these developments, GR has been extended to multi-modal tasks \cite{GR_MM, GR_MM1}. GRACE \cite{GR_IR_GRACE} introduces structured natural language identifiers to enable generative cross-modal retrieval within MLLMs \cite{InternVideo2.5}. IRGen \cite{GR_IR_IRGEN} employs discrete visual tokens for image retrieval, while AVG \cite{GR_IR_AVG} introduces cross-modal alignment during tokenization.
Video retrieval presents unique challenges due to the semantic richness of videos. T2VIndexer \cite{GT_TVR_T2VIndexer} extends GR to video retrieval by constructing hierarchical video identifiers through offline clustering. However, videos are inherently polysemous, yet these approaches assign each video a single semantic ID. This one-to-one mapping forces diverse query intents to converge on identical codes. 

\begin{figure*}
    \centering
    \includegraphics[width=1\linewidth]{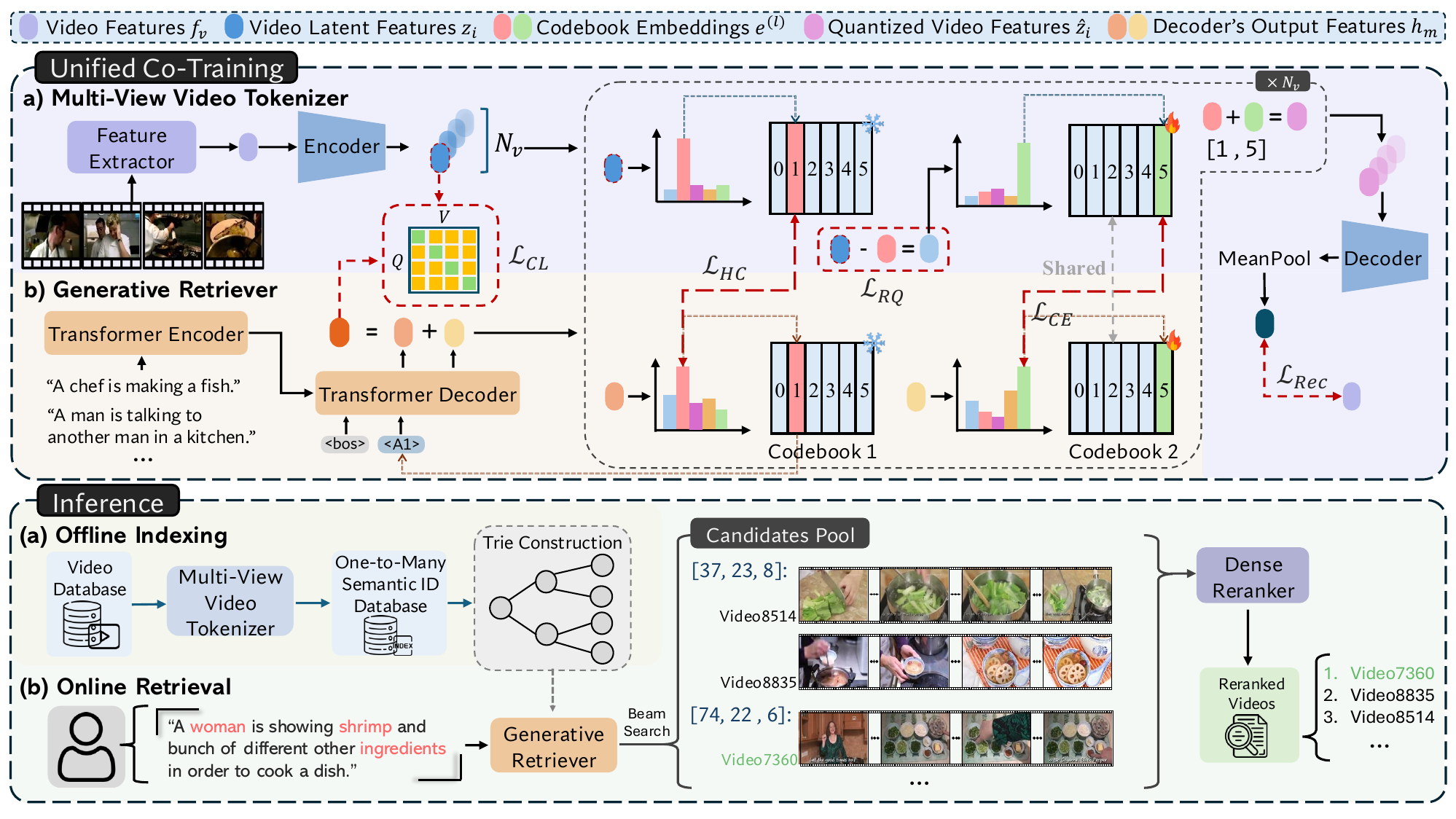}
    \vspace{-15pt}
    \caption{Overview of the GRDR framework. Training: (a) The multi-view video tokenizer encodes videos into discrete semantic IDs. It employs cross-modal alignment ($\mathcal{L}_{CL}$) to align video latent features $z_{i}$ with the retriever’s cumulative decoder features $h^{(m)}$ via contrastive learning. The video latent features $z_i$ then passed to residual quantization ($\mathcal{L}_{RQ}$), followed by a reconstruction decoder ($\mathcal{L}_{Rec}$) to prevent semantic collapse. (b) The generative retriever is jointly optimized with the tokenizer via a shared codebook. Its decoder predicts codes using cosine similarity, trained with cross-entropy loss ($\mathcal{L}_{CE}$). We employ progressive training, where Hierarchical Consistency ($\mathcal{L}_{HC}$) loss is used to prevent code drift in previous layers. The inference pipeline follows a Generative Recall, Dense Reranking paradigm: (a) Offline Indexing: Videos are tokenized into compact IDs to construct a prefix Trie for constrained decoding. (b) Online Retrieval: The retriever generates candidate Semantic IDs via beam search.}
    \label{fig:training}   
    \vspace{-10pt}
\end{figure*}

\section{Methodology}

\subsection{Preliminaries}




For a given text query $q$, TVR seeks to identify the most semantically relevant video(s) $v^*$ from a corpus $\mathcal{V}=\{v_i\}_{i=1}^N$. In the generative retrieval paradigm, the retrieval process is formulated as a sequence-to-sequence generation problem. This paradigm comprises two core components: (i) a video tokenizer that quantizes continuous video representations into discrete semantic identifiers, and (ii) a generative retriever that auto-regressively predicts the target identifier conditioned on the query. We provide a brief formulation of each component below.

\noindent \textbf{Video Tokenizer.} The video tokenizer $\mathcal{T}(\cdot)$ maps each video $v$ into a  sequence of $m$ discrete tokens, which we term a semantic ID: 
\begin{equation}
    \text{semantic-ID}_v = \mathcal{T}(v) = \{c_j\}_{j=1}^m,
\end{equation}
where $c_j$ ($j=1,2,\ldots,m$) is the $j$-th discrete token in the sequence.



\noindent \textbf{Generative Retriever.} With the semantic IDs assigned, a generative retriever learns a probabilistic model $\Theta(\cdot)$ (e.g. T5 model \cite{T5}) that maximizes the likelihood of generating the target semantic ID given a text query $q$, where the sequence of tokens in a semantic ID is generated in an auto-regressive manner by maximizing the joint probability:
\begin{equation}
    \text{semantic-ID}_v = \{c_j\}_{j=1}^{m} = \Theta(q), \,\,\,\,\text{s.t.}\,\, \max\prod_{j=1}^m P(c_j \mid q, c_{<j}, \Theta),
\end{equation}

where $c_{<j}$ denotes the prefix of the semantic ID generated prior to step $j$. 
During inference, through beam search \cite{Transformer}, $\Theta(\cdot)$ retrieves the most relevant video(s) for query $q$ by auto-regressively decoding $q$ into corresponding semantic ID(s) with maximum joint likelihood.

\subsection{Multi-View Video Tokenizer}
\label{sec:video_tokenizer}

Videos are inherently polysemous, yet existing video tokenizers compress each video into a single semantic ID. This one-to-one mapping introduces semantic ambiguity: recall fails when the generated path misses a valid semantic facet. To address this, each video requires parallel retrieval paths. This design allows matching any single path suffices to recall the video, improving coverage across diverse query intents. This motivates our multi-view video tokenizer, which assigns multiple semantic IDs per video. In the meantime, directly training multiple video encoders is infeasible as it risks two failures. First, views tend to converge to identical representations due to representation collapse. Furthermore, diversity must align with actual query intents, not arbitrary decompositions.



To counter those limitations, we design a multi-view video tokenizer that ensures both diversity and coherence. This design enables parallel retrieval paths for each video. As shown in Figure \ref{fig:training}(a), our tokenizer comprises three components: (i) a query-guided multi-view video encoder that projects video features into query-adaptive latent representations, enforcing diversity aligned with real query intents, (ii) a residual quantizer discretizes each view into a hierarchical code sequence, and (iii) a reconstruction decoder that ensures semantic coherence.




\noindent \textbf{Query-Guided Multi-View Video Encoder.} The multi-view video encoder maps each video into $N_v$ distinct latent features, where each feature captures a distinct semantic view. Specifically, given a raw video $v$ with $T$ sampled frames, we first extract video representations using a pretrained Vision-Language Model (VLM) (e.g., InternVideo2 \cite{InternVideo2}) as backbone. This shared video feature then serves as input to $N_v$ distinct encoder networks $\{\phi_1, \phi_2, ..., \phi_{N_v}\}$ :

\begin{equation}
\label{video_latent_features}
    \mathbf{z}_i = \phi_i(\mathbf{f}_v), \quad i \in \{1, 2, ..., N_v\},
\end{equation}
where $\mathbf{z}_i \in \mathbb{R}^{d_z}$. This encoder design enables one video to be retrieved through multiple distinct paths.

To ensure each view captures semantically distinct features without collapse, we guide the view encoders with query intent supervision. For each video, we extract its associated query features using the same VLM \cite{InternVideo2} backbone. We then cluster these features via k-means with $K = N_v$ clusters. Each cluster represents a distinct semantic intent that users may express when searching for the video. During training, queries are hard-assigned to clusters, and sub-encoder $\phi_i$ learns to align with queries in cluster $i$. We enforce this alignment through a cross-modal contrastive loss $\mathcal{L}_{CL}$ between each view's latent feature $\mathbf{z}_i$ and the corresponding query features extracted from the generative retriever's decoder (Eq. \ref{eq:CL_loss}). Through query-guided encoding, we ground diversity in real retrieval patterns rather than arbitrary decompositions.

\noindent \textbf{Residual Quantization.} Each view's latent features $\mathbf{z}_i$ are discretized into a semantic ID sequence through Residual Quantization (RQ) \cite{RQ}. RQ employs $M$ codebook layers, where each codebook $\mathcal{C}^{(m)} = \{\mathbf{e}_1^{(m)}, \dots, \mathbf{e}_K^{(m)}\}$ contains $K$ learnable embedding vectors. These codebooks serve as the discrete vocabulary for quantization, and the selected code indices form the final semantic ID. The quantization proceeds layer-by-layer through iterative refinement. At each layer $m$, we first compute the residual between the latent features and its accumulated approximation from previous layers:
\begin{equation}
\label{residual_features}
    \mathbf{r}^{(1)}_i = \mathbf{z}_i, \quad \mathbf{r}^{(m)}_i = \mathbf{z}_i - \sum_{l=1}^{m-1} \mathbf{e}^{(l)}_{c^{(l)}_i}, \quad m > 1.
\end{equation}
We then select the codebook entry with maximum cosine similarity to the current residual:
\begin{equation}
\label{code_selection}
    c^{(m)}_i = \underset{k \in \{1,\dots,K\}}{\arg\max} \ \cos(\mathbf{r}^{(m)}_i, \mathbf{e}^{(m)}_k),
\end{equation}
where we adopt cosine similarity $\cos(\cdot)$ instead of $L_2$ distance for code selection to enable cross-modal alignment with the generative retriever, as detailed in Section \ref{sec:uni-co}. 

We minimize the quantization error of RQ through two complementary loss terms:
\begin{equation}
    \mathcal{L}_{RQ} = \underbrace{\|\hat{\mathbf{z}}_i - \operatorname{sg}[\mathbf{z}_i]\|_2^2}_{\text{codebook loss}} + \beta \cdot \underbrace{\|\operatorname{sg}[\hat{\mathbf{z}}_i] - \mathbf{z}_i\|_2^2}_{\text{commitment loss}},
\end{equation}
where $\operatorname{sg}[\cdot]$ denotes the stop-gradient operator and $\hat{\mathbf{z}}_i$ denotes the reconstructed quantized video features by summing the selected embeddings.

\noindent \textbf{Reconstruction Decoder.} We leverage a reconstruction decoder to preserve video semantics and prevent cross-view collapse during quantization. The design of decoder mirrors the multi-view encoder with $N_v$ view-specific sub-decoders $\{\psi_i\}_{i=1}^{N_v}$. Each sub-decoder maps the quantized view-specific features $\hat{\mathbf{z}}_i$ into reconstructed features:
\begin{equation}
\label{reconstructed_features}
    \tilde{\mathbf{f}}_i = \psi_i(\hat{\mathbf{z}}_i), \quad i \in \{1, 2, ..., N_v\}.
\end{equation}
The final reconstructed video features $\tilde{\mathbf{f}}_v$ is obtained via mean pooling across all views. With $\tilde{\mathbf{f}}_v$, we utilize a cosine distance reconstruction loss $\mathcal{L}_{Rec} = 1 - \cos(\mathbf{f}_v, \tilde{\mathbf{f}}_v)$ on top of the code selection objective $\mathcal{L}_{RQ}$ to regularize the quantization.

\subsection{Unified Co-Training}
\label{sec:uni-co}

Cross-modal misalignment is a critical obstacle to applying GR as an effective recall stage for TVR.  As the video tokenizer and the generative retriever are optimized in two isolated stages, retrieval supervision cannot propagate back to the video tokenization process.
A natural solution is to inject the cross-modal alignment objective during the tokenizer training stage (\textit{e.g.,} AVG \cite{GR_IR_AVG}), but this remains a two-stage design, disconnecting the optimization of semantic IDs with text-based retrieval. 

This motivates our unified co-training that couples the tokenizer with the retriever, such that retrieval-specific signals directly refine semantic ID assignment. However, the tokenizer and retriever rely on incompatible code selection geometries, \textit{i.e.,} tokenizer nearest-neighbor assignment under $L_2$ distance, while the retriever predicts codes via dot-product with softmax. Therefore, straightforwardly aligning probability distributions between them for co-training can cause training collapse and degrade performance. 

As per Figure \ref{fig:training}(b), we achieve unified co-training through:  (i) a shared codebook couples the tokenizer and retriever, enabling bidirectional gradient flow via a unified code selection strategy, and (ii) a progressive training strategy that jointly refines all components based on retrieval objectives. We describe the design below.

\renewcommand{\algorithmicrequire}{\textbf{Input:}}
\begin{algorithm}[t]
\caption{Pseudocode of Progressive Co-Training for GRDR}
\label{alg:progressive_training}
\begin{algorithmic}[1]
\small
\REQUIRE $\mathbf{f}_v$, $q$, codebook layers $M$, size $K$
\FOR{$m = 1$ \textbf{to} $M$}
    \STATE Freeze $\mathcal{C}^{(1:m-1)}$
    { \renewcommand{\algorithmicdo}{\textbf{do} \textcolor{gray}{$\triangleright$ P1: Cross-modal alignment}} 
    \FOR{\texttt{epoch} $= 1$ \textbf{to} \texttt{align\_epochs}} 
        \renewcommand{\algorithmicdo}{\textbf{do}} 
        
        \FOR{\textbf{each} $(v, q)$ \textbf{in} $\mathcal{D}$}
            \STATE $\mathbf{z}_i \leftarrow$ Eq.~(\ref{video_latent_features}); \quad $\mathbf{h}^{(m)} \leftarrow \sum_{l=1}^{m} \mathbf{h}_l$
            \STATE $\mathcal{L} \leftarrow$ Eq.~(\ref{eq:CL_loss}) $+ \; \mathbbm{1}_{m>1} \cdot$ Eq.~(\ref{HC_loss})
            \STATE Back-propagate and update
        \ENDFOR
    \ENDFOR }
    \STATE $\mathbf{r}^{(m)}_i \leftarrow$ Eq.~(\ref{residual_features}); \!\!\! \quad $\mathcal{C}^{(m)} \leftarrow \texttt{KMeans}(\mathbf{r}^{(m)}, K)$ \textcolor{gray}{$\triangleright$ Initialize codebook}
    { \renewcommand{\algorithmicdo}{\textbf{do} \textcolor{gray}{$\triangleright$ P2: Unified co-training}} 
    \FOR{\texttt{epoch} $= 1$ \textbf{to} \texttt{train\_epochs}}
        \renewcommand{\algorithmicdo}{\textbf{do}} 
        \FOR{\textbf{each} $(v, q)$ \textbf{in} $\mathcal{D}$}
            \STATE $c^{(m)}_i \leftarrow$ Eq.~(\ref{code_selection}); \quad $\tilde{\mathbf{f}}_v \leftarrow$ Eq.~(\ref{reconstructed_features}) \textcolor{gray}{$\triangleright$ Video tokenizer}
            \STATE $P(c_m|c_{<m}, q) \leftarrow$ Eq.~(\ref{likelyhood_prediction}) \textcolor{gray}{$\triangleright$ Soft prediction from retriever}
            \STATE $\mathcal{L}_{total} \leftarrow$ Eq.~(\ref{total_objective})
            \STATE Back-propagate and update
        \ENDFOR
    \ENDFOR }
\ENDFOR
\end{algorithmic}
\end{algorithm}


\noindent \textbf{Shared Codebook Mechanism.} The codebook serves as both the quantization vocabulary in the video tokenizer and the generation vocabulary in the generative retriever. To couple both components into end-to-end manner, we use the same codebook $C^{(m)}$ for both video tokenization and semantic IDs generation at each layer $m$. In the generative retriever, we compute token probabilities by applying a softmax over the cosine similarities between the retriever’s decoder output features $h_m$ and the codebook embeddings $C^{(m)}$:
\begin{equation}
\label{likelyhood_prediction}
    P(c_m | c_{<m}, q) = \text{softmax}\left(\cos(\mathbf{h}_m, \mathbf{C}^{(m)}) / \tau \right).
\end{equation}
Meanwhile, the tokenizer assigns codes by selecting the nearest codebook entry under cosine similarity over the same $C^{(m)}$ (Eq. \ref{code_selection}). Gradients from the retrieval loss can directly update the codebook, which in turn refines the tokenizer’s quantization boundaries. Further, since both components operate on cosine similarity with shared embeddings, both modalities can be optimized in a shared hypersphere space.

This unified code selection strategy allows us to align video and text features before codebook learning. A contrastive learning objective ($\mathcal{L}_{CL}$) directly aligns the tokenizer's latent features $\mathbf{z}_i$ with the retriever's cumulative decoder features $\mathbf{h}^{(m)} = \sum_{l=1}^{m} \mathbf{h}_l$, mirroring the residual addition structure in video quantization:
\begin{equation}
    \mathcal{L}_{CL} = -\log \frac{\exp(\cos(\mathbf{z}_i, \mathbf{h}^{(m)}) / \tau)}{\sum_{j=1}^{B} \exp(\cos(\mathbf{z}_j, \mathbf{h}^{(m)}) / \tau)},
\label{eq:CL_loss}
\end{equation}
where $\tau$ is a learnable temperature parameter. This objective pulls matched video-text pairs together in the shared embedding space. In addition, when optimizing codebook for layer $m > 1$, we apply a hierarchical consistency (HC) loss $\mathcal{L}_{HC}$ on both modalities to prevent the learned codes in all $m-1$ previous layers from drifting:
\begin{equation}
\label{HC_loss}
    \mathcal{L}_{HC} = -\sum_{l=1}^{m-1} \left[ \log P(c_l | c_{<l}, q) + \log P(c_l | c_{<l}, v) \right].
\end{equation}
This loss ensures both the retriever and tokenizer retain prediction capability for codes from preceding layers.

\noindent \textbf{Progressive Training Strategy.} The hierarchical structure of residual quantization naturally motivates a progressive training strategy. Training all $M$ layers simultaneously risks unstable optimization and suboptimal code allocation. We instead train the framework progressively, one codebook layer at a time. As illustrated in Algorithm 
\ref{alg:progressive_training}, the training pipeline of each layer is split into two phases. We first align both modalities into a shared embedding space to allow query-guided video encoding. After alignment, we initialize codebook $\mathbf{C}^{(m)}$ via k-means clustering on residual features, providing semantically meaningful starting points. We then jointly optimize both components. The retriever learns to generate target semantic IDs through a cross-entropy loss:
\begin{equation}
    \mathcal{L}_{CE} = -\log P(c_m | c_{<m}, q),
\end{equation}
where $c_m$ denotes the code assigned by Video Tokenizer at layer $m$. The full co-training objective combines all losses:
\begin{equation}
\label{total_objective}
    \mathcal{L}_{total}= \underbrace{\lambda_1 \mathcal{L}_{CE} + \lambda_2 \mathcal{L}_{HC}}_{\text{Generative Retriever}} + \underbrace{\lambda_3 \mathcal{L}_{RQ} + \lambda_4 \mathcal{L}_{Rec}}_{\text{Video Tokenizer}},
\end{equation}
where $\lambda_1, ..., \lambda_4$ are weighting coefficients.

\begin{table*}[t]
\definecolor{lightyellow}{RGB}{255, 249, 192}
\definecolor{lightpurple}{RGB}{248, 231, 242}
\definecolor{timecolor}{RGB}{0, 115, 140}
\newcommand{\lat}[1]{\textcolor{timecolor}{#1}}
\centering
\caption{Performance comparison on standard inductive setting. The search pool is restricted to the test set. The top two Generative Recall + Dense Rerank models are highlighted in bold and underlined. \lat{$T_{\text{latency}}$} is measured in milliseconds (ms).}
\vspace{-5pt}
\label{tab:Inductive}
\setlength{\aboverulesep}{0pt}
\setlength{\belowrulesep}{0pt}
\resizebox{\textwidth}{!}{%
\begin{tabular}{l c ccc c ccc c ccc c ccc c}
\toprule
\multicolumn{1}{l}{\multirow{2}{*}{\textbf{Method}}} & \multirow{2}{*}{\textbf{Params}} & \multicolumn{4}{c}{\textbf{MSR-VTT} \cite{MSRVTT}} & \multicolumn{4}{c}{\textbf{ActivityNet} \cite{ACTNET}} & \multicolumn{4}{c}{\textbf{DiDeMo} \cite{DiDeMo}} & \multicolumn{4}{c}{\textbf{LSMDC} \cite{LSMDC}} \\
\cmidrule(lr){3-6} \cmidrule(lr){7-10} \cmidrule(lr){11-14} \cmidrule(lr){15-18}
 & & R@1 & R@5 & R@10 & \lat{$T_{\text{latency}}$} & R@1 & R@5 & R@10 & \lat{$T_{\text{latency}}$} & R@1 & R@5 & R@10 & \lat{$T_{\text{latency}}$} & R@1 & R@5 & R@10 & \lat{$T_{\text{latency}}$} \\
\midrule
\multicolumn{2}{l}{\textbf{Search Pool Size}} & \multicolumn{4}{c}{1,000} & \multicolumn{4}{c}{4,917} & \multicolumn{4}{c}{1,003} & \multicolumn{4}{c}{1,000} \\
\midrule
\rowcolor{lightyellow}
\multicolumn{18}{l}{\textit{\textbf{Dense Retrieval}}} \\
CLIP4CLIP \cite{TVR_CLIP4clip} & 151M & 44.5 & 71.4 & 81.6 & \lat{251} & 40.5 & 72.4 & - & \lat{1211} & 43.4 & 70.2 & 80.6 & \lat{245} & 22.6 & 41.0 & 49.1 & \lat{243} \\
X-Pool \cite{TVR_Xpool} & 153M & 46.9 & 72.8 & 82.2 & \lat{388} & 35.0 & 65.9 & 78.5 & \lat{1899} & 41.4 & 70.6 & 79.9 & \lat{394} & 25.2 & 43.7 & 53.5 & \lat{382} \\
InternVideo2-1B \cite{InternVideo2} & 1412M & 45.9 & 70.6 & 78.7 & \lat{255} & 47.4 & 76.0 & 85.2 & \lat{1214} & 45.3 & 71.6 & 79.6 & \lat{257} & 25.1 & 42.7 & 50.5 & \lat{255} \\
\midrule
\rowcolor{lightyellow}
\multicolumn{18}{l}{\textit{\textbf{Dense Recall + Dense Rerank}}} \\
EERCF \cite{TVR_EERCF} & 164M & 47.8 & 74.1 & 84.1 & \lat{2638} & 43.1 & 74.5 & 86.0 & \lat{5179} & 27.1 & 54.5 & 63.8 & \lat{2292} & 17.0 & 32.1 & 41.1 & \lat{2554} \\
\midrule
\rowcolor{lightyellow}
\multicolumn{18}{l}{\textit{\textbf{Generative Recall + Dense Rerank \cite{TVR_Xpool}}}} \\
TIGER \cite{GR_RecSys_TIGER} & 69M & 36.0 & 55.3 & 62.5 & \lat{245} & 8.9 & 15.9 & 18.3 & \lat{243} & 17.4 & 28.9 & 32.8 & \lat{242} & 4.6 & 8.3 & 10.1 & \lat{264} \\
AVG \cite{GR_IR_AVG} & 69M & 40.4 & \underline{61.4} & \underline{69.9} & \lat{227} & 14.3 & 26.4 & 30.5 & \lat{237} & 24.9 & 40.5 & 45.3 & \lat{235} & 11.7 & 18.7 & 21.3 & \lat{234} \\
T2VIndexer \cite{GT_TVR_T2VIndexer} & 144M & \underline{42.3} & \underline{61.4} & 69.6 & \underline{\lat{179}} & \underline{26.0} & \underline{46.4} & \underline{53.9} & \underline{\lat{183}} & \underline{34.8} & \underline{53.4} & \underline{60.4} & \underline{\lat{179}} & \underline{20.6} & \underline{33.7} & \underline{38.2} & \underline{\lat{193}} \\
\midrule
\rowcolor{lightpurple}
\textbf{GRDR (Ours)} & \textbf{63M} & \textbf{46.0} & \textbf{70.1} & \textbf{78.0} & \textbf{\lat{136}} & \textbf{33.7} & \textbf{63.7} & \textbf{76.6} & \textbf{\lat{125}} & \textbf{39.9} & \textbf{65.8} & \textbf{74.2} & \textbf{\lat{118}} & \textbf{23.5} & \textbf{39.4} & \textbf{46.2} & \textbf{\lat{144}} \\
\bottomrule
\end{tabular}%
}
\end{table*}

\subsection{Inference}
\label{sec:inference}

As shown in Figure \ref{fig:training}, the inference pipeline of GRDR formalizes the generative recall followed by dense rerank. Given a text query, the generative retriever first decodes a small set of semantic ID candidates, and a dense TVR model then reranks the candidate videos to recover fine-grained matching accuracy. 
This inference pipeline decouples into offline indexing and online retrieval phases, which achieves two efficiency gains: (i) compact storage through discrete code representation, and (ii) corpus-size-independent retrieval through constrained auto-regressive decoding.

\noindent \textbf{Offline Indexing.} Given a video corpus $V$, video tokenizer encodes each video into $N_v$ semantic IDs. All the semantic IDs in the corpus are organized into a prefix trie, which constrains the generative decoding space to valid semantic IDs. 

\noindent \textbf{Online Retrieval.} At query time, the generative retriever processes the input text and generates semantic ID candidates via trie-constrained beam search. 
The generated semantic IDs are then mapped to video candidates via the database. Given that each video maps to $N_v$ semantic IDs and multiple videos may share the same code, we de-duplicate the retrieved candidates while preserving the beam ranking order. Finally, the candidate set is reranked by a dense retrieval model. 

\begin{table*}[t]
\definecolor{lightyellow}{RGB}{255, 249, 192}
\definecolor{lightpurple}{RGB}{248, 231, 242}
\definecolor{timecolor}{RGB}{0, 115, 140}
\newcommand{\lat}[1]{\textcolor{timecolor}{#1}}

\centering
\caption{Performance comparison on the full-corpus setting. The search pool is the union of training and test sets. The top two Generative Recall + Dense Rerank models are highlighted in \textbf{bold} and \underline{underlined}. \lat{$T_{\text{latency}}$} is measured in milliseconds (ms).}
\vspace{-5pt}
\label{tab:full_corpus}
\setlength{\aboverulesep}{0pt}
\setlength{\belowrulesep}{0pt}
\resizebox{\textwidth}{!}{%
\begin{tabular}{l c ccc c ccc c ccc c ccc c}
\toprule
\multicolumn{1}{l}{\multirow{2}{*}{\textbf{Method}}} & \multirow{2}{*}{\textbf{Params}} & \multicolumn{4}{c}{\textbf{MSR-VTT} \cite{MSRVTT}} & \multicolumn{4}{c}{\textbf{ActivityNet} \cite{ACTNET}} & \multicolumn{4}{c}{\textbf{DiDeMo} \cite{DiDeMo}} & \multicolumn{4}{c}{\textbf{LSMDC} \cite{LSMDC}} \\
\cmidrule(lr){3-6} \cmidrule(lr){7-10} \cmidrule(lr){11-14} \cmidrule(lr){15-18}
& & R@1 & R@5 & R@10 & \lat{$T_{\text{latency}}$} & R@1 & R@5 & R@10 & \lat{$T_{\text{latency}}$} & R@1 & R@5 & R@10 & \lat{$T_{\text{latency}}$} & R@1 & R@5 & R@10 & \lat{$T_{\text{latency}}$} \\
\midrule
\multicolumn{2}{l}{\textbf{Search Pool Size}} & \multicolumn{4}{c}{10,000} & \multicolumn{4}{c}{14,926} & \multicolumn{4}{c}{9,384} & \multicolumn{4}{c}{102,055} \\
\midrule
\rowcolor{lightyellow}
\multicolumn{18}{l}{\textit{\textbf{Dense Retrieval}}} \\
CLIP4CLIP \cite{TVR_CLIP4clip} & 151M & 10.8 & 21.7 & 28.1 & \lat{2456} & 20.5 & 43.4 & 55.6 & \lat{3658} & 19.0 & 36.9 & 48.2 & \lat{2282} & 4.0 & 7.6 & 10.6 & \lat{24941} \\
X-Pool \cite{TVR_Xpool} & 153M & 22.0 & 41.0 & 51.2 & \lat{3795} & 21.5 & 45.7 & 57.9 & \lat{5641} & 22.5 & 42.6 & 51.6 & \lat{3726} & 5.4 & 10.2 & 12.1 & \lat{38029} \\
InternVideo2-1B \cite{InternVideo2} & 1412M & 24.4 & 43.2 & 52.4 & \lat{2460} & 33.9 & 59.5 & 70.9 & \lat{3661} & 26.0 & 46.3 & 54.5 & \lat{2286} & 5.7 & 10.4 & 13.9 & \lat{24945} \\
\midrule
\rowcolor{lightyellow}
\multicolumn{18}{l}{\textit{\textbf{Dense Recall +  Dense Rerank}}} \\
EERCF \cite{TVR_EERCF} & 164M & 20.5 & 42.7 & 50.9 & \lat{4843} & 18.9 & 41.3 & 53.4 & \lat{7628} & 10.4 & 27.2 & 35.3 & \lat{4329} & 3.4 & 6.4 & 8.8 & \lat{27252} \\
\midrule
\rowcolor{lightyellow}
\multicolumn{18}{l}{\textit{\textbf{Generative Recall + Dense Rerank \cite{TVR_Xpool}}}} \\
TIGER \cite{GR_RecSys_TIGER} & 69M & 8.2 & 14.6 & 18.4 & \lat{218} & 3.7 & 6.6 & 8.1 & \lat{248} & 3.9 & 6.6 & 7.7 & \lat{243} & 0.1 & 0.1 & 0.1 & \lat{553} \\
AVG \cite{GR_IR_AVG} & 69M & 10.3 & 19.8 & 24.3 & \lat{221} & 4.9 & 9.9 & 12.4 & \lat{219} & 4.5 & 7.5 & 8.7 & \lat{220} & 0.2 & 0.4 & 0.4 & \lat{235} \\
T2VIndexer \cite{GT_TVR_T2VIndexer} & 144M & \underline{16.0} & \underline{31.2} & \underline{38.8} & \underline{\lat{198}} & \underline{17.5} & \underline{37.9} & \underline{49.0} & \underline{\lat{207}} & \underline{13.0} & \underline{23.9} & \underline{28.9} & \underline{\lat{205}} & \underline{1.2} & \underline{1.9} & \underline{2.7} & \underline{\lat{211}} \\
\rowcolor{lightpurple}
\textbf{GRDR (Ours)} & \textbf{63M} & \textbf{17.4} & \textbf{32.2} & \textbf{39.7} & \textbf{\lat{184}} & \textbf{19.2} & \textbf{41.1} & \textbf{51.8} & \textbf{\lat{116}} & \textbf{15.5} & \textbf{29.7} & \textbf{36.1} & \textbf{\lat{119}} & \textbf{2.1} & \textbf{4.8} & \textbf{5.9} & \textbf{\lat{121}} \\
\bottomrule
\end{tabular}%
}
\end{table*}

\section{Experiments}
To validate our contributions and demonstrate the practical benefits of GRDR, we conduct extensive experiments to answer the following research questions (RQs): \textbf{(RQ1)} How effective is GRDR compared to existing TVR baselines? \textbf{(RQ2)} How much index storage and query-time latency does GRDR save compared to baselines during practical corpus-scale retrieval? \textbf{(RQ3)} How do the main architectural components of GRDR contribute to its retrieval performance? \textbf{(RQ4)} How do key hyperparameters of GRDR affect its retrieval performance and query-time latency?

\subsection{Experiment Setup}
\noindent \textbf{Datasets.}
We evaluate GRDR on four widely used text-to-video retrieval benchmarks. \textbf{MSR-VTT} \cite{MSRVTT} consists of 10,000 video clips from 20 categories, each annotated with 20 sentences. We utilize the standard split of 9,000 videos for training and 1,000 for testing. \textbf{ActivityNet} \cite{ACTNET} comprises 19,000 YouTube videos across 200 activity categories, averaging 5 captions per clip. We utilize the standard partition of 10,009 training and 4,917 test samples. \textbf{LSMDC} \cite{LSMDC} contains 118,000 short clips derived from 202 movies, providing 101,055 training and 1,000 test samples. Finally, \textbf{DiDeMo} \cite{DiDeMo} consists of 10,000 videos with 26,000 detailed moment descriptions, averaging 5 captions per clip. We utilize the split with 8,381 training and 1,003 test samples.

\noindent \textbf{Evaluation Metrics.} 
We evaluate retrieval performance using standard Recall@K metrics (R@1, R@5, R@10). We measure the average inference latency per query to evaluate retrieval efficiency. 
For two-stage methods, the total inference latency per query $T_{\text{latency}}$ combines both recall and rerank stages: $T_{\text{latency}} = T_{\text{recall}} + T_{\text{rerank}}$, where in cases of a GR-based recall model, $T_{\text{recall}}$ denotes the semantic ID generation time via beam search, including query encoding and auto-regressive decoding. We report $T_{\text{latency}}$ by averaging across the test queries. This metric directly reflects the system's response time in deployment scenarios.

\noindent \textbf{Evaluation Settings.} We evaluate all methods under two distinct settings to test different capabilities:

\noindent \textbf{(i) Inductive Setting.} Following the standard protocol in prior TVR work \cite{TVR_CLIP4clip, TVR_Xpool, TVR_EERCF}, the search pool in the inductive setting is restricted to unseen videos from the test set. This setting evaluates the model's ability to generalize to unseen videos. 

\noindent \textbf{(ii) Full-Corpus Setting.} The search pool contains the union of training and test videos. We introduce this setting to better reflect real-world deployment, where systems must locate content within dynamic libraries containing both historical (seen) and newly uploaded (unseen) videos. 

\vspace{-1em}
\subsection{Baselines}
\noindent \textbf{Dense Retrieval Methods.}
We compare against representative dense retrieval baselines. \textbf{CLIP4CLIP} \cite{TVR_CLIP4clip} adapts dual-encoder architecture to video retrieval with temporal pooling. \textbf{X-Pool} \cite{TVR_Xpool} extends this with frame-level cross-modal attention. \textbf{InternVideo2} \cite{InternVideo2} scales to 1B parameters with large-scale pretraining, achieving SOTA result in video tasks. We use it as the backbone for all GR methods in our experiments. \textbf{EERCF} \cite{TVR_EERCF} is the only dense method adopting the recall-and-rerank paradigm, using dense recall followed by frame-level and patch-level interaction for reranking. We integrate the official implementations of these methods into our GRDR codebase.

\noindent \textbf{Generative Retrieval Methods.}
GR methods serve as effective candidate generators during the recall stage. \textbf{TIGER} \cite{GR_RecSys_TIGER} first establishes GR for sequential recommendation, demonstrating strong generative retrieval capability. \textbf{AVG} \cite{GR_IR_AVG} extends the TIGER to image retrieval by incorporating cross-modal alignment during tokenization. The tokenizer leverages pre-trained video features to produce semantic IDs that better capture visual-textual alignment. \textbf{T2VIndexer} \cite{GT_TVR_T2VIndexer} is tailored for video retrieval by constructing hierarchical semantic IDs through semantic tree clustering. For fair comparison, we utilize the same InternVideo2-Stage2-1b~\cite{InternVideo2} as video feature encoder and T5-Small \cite{T5} as the retriever, and plug in the state-of-the-art dense model X-Pool \cite{TVR_Xpool} to perform reranking. We configure all baseline methods with 4 codebook layers of size 256 to ensure sufficient semantic capacity. 
\subsection{Implementation Details}

\textbf{Model Architecture.} 
We extract video features $f_v$ using the \\ InternVideo2-Stage2-1b \cite{InternVideo2} backbone. We configure the Multi-View Video Tokenizer with $N_v = 4$, codebook layers $M = 3$, and length $K = 128$. For LSMDC \cite{LSMDC}, we adjust these to $N_v = 1$ and $K = 200$ to accommodate its specific domain and scale. For consistency with other baselines, we employ T5-Small \cite{T5} as the generative retriever and X-Pool \cite{TVR_Xpool} for dense reranking.

\noindent \textbf{Training Configuration.} 
We use AdamW optimizer \cite{AdamW} with learning rate $1 \times 10^{-4}$ and batch size 512. We employ a progressive training schedule: The first codebook layer requires 3 epochs of cross-modal pre-alignment and 4 of co-training. Subsequent layers reduce alignment to 1 epoch while maintaining 4 epochs of co-training. To provide sufficient query diversity for multi-view learning, we generate pseudo queries using GPT-5-mini \cite{noauthor_gpt-5_nodate}. Each video is augmented to around 50 queries for MSR-VTT \cite{MSRVTT}, ActivityNet \cite{ACTNET}, and DiDeMo \cite{DiDeMo}.

\noindent \textbf{Inference Configuration.} 
Beam size $B$ is tuned per dataset to generate approximately 100-150 candidate videos per query, balancing recall coverage against reranking cost. Generated semantic IDs are then mapped to videos and deduplicated. During deduplication, we preserve the beam ranking order to maintain retrieval priority.

\noindent \textbf{Computational Resources.} 
All experiments are conducted on a single NVIDIA A6000 GPU (48GB). For latency measurement, we simulate real-time single-query scenarios using a batch size of 1, reporting the average over at least 100 queries after warm-up.

\subsection{Overall Performance (RQ1)}
\noindent \textbf{Inductive Setting.} Table \ref{tab:Inductive} shows GRDR consistently outperforms all GR baselines across four benchmarks. Since X-Pool~\cite{TVR_Xpool} serves as our reranker, we use its standalone performance as the effectiveness ceiling. GRDR approaches this ceiling with minimal gaps of 0.9 on MSR-VTT \cite{MSRVTT}. These minimal differences indicate our recall stage successfully retrieves nearly all videos the reranker needs for accurate ranking. In contrast, TIGER \cite{GR_RecSys_TIGER} and AVG \cite{GR_IR_AVG} suffer from semantic ambiguity due to one-to-one mapping. Although AVG \cite{GR_IR_AVG} improves over TIGER \cite{GR_RecSys_TIGER} by 4.4 on MSR-VTT \cite{MSRVTT} through text features incorporation, it still remains 5.6 below GRDR. T2VIndexer's \cite{GT_TVR_T2VIndexer} offline clustering without retrieval feedback produces codes misaligned with textual relevance, making its performance consistently below GRDR. EERCF \cite{TVR_EERCF} shows degraded performance on some dataset due to our computational constraints. Despite this, it remains valuable as the sole Dense Recall + Dense Rerank baseline for efficiency comparison.

\noindent \textbf{Full-Corpus Setting.} The expanded search pool amplifies scalability challenges as training videos create difficult distractors with high semantic overlap. As shown in Table \ref{tab:full_corpus}, all methods suffer substantial degradation. However, GRDR maintains superior performance over GR baselines. Notably, TIGER \cite{GR_RecSys_TIGER} and AVG \cite{GR_IR_AVG} collapse to 0.1 and 0.2 R@1 on LSMDC \cite{LSMDC}. This almost failed performance demonstrates that semantic ambiguity becomes critical at scale, where a single semantic ID cannot disambiguate nuanced content. Interestingly, among dense retrieval methods, InternVideo2 \cite{InternVideo2} outperforms all others even on datasets where it underperformed in the Inductive Setting. This suggests that model scaling improves generalization to larger corpora.

\begin{figure}
    \centering
    \includegraphics[width=1\linewidth]{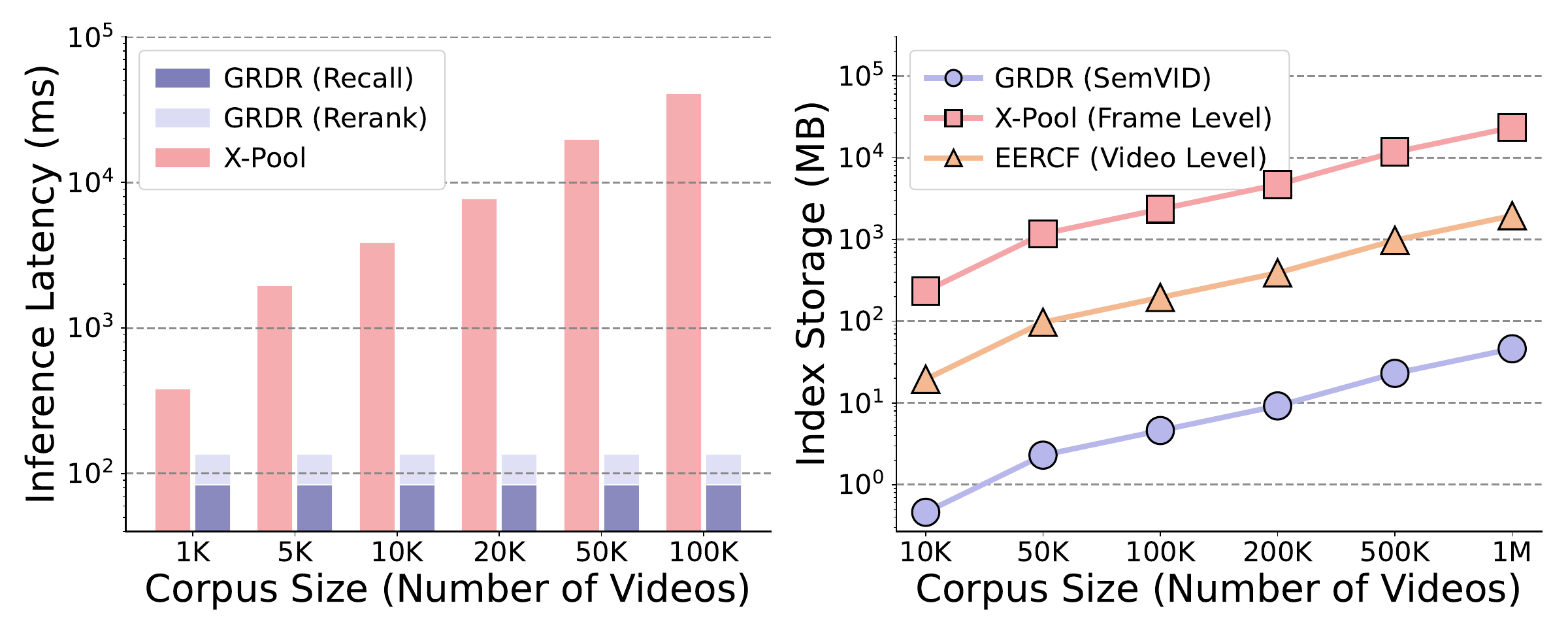}
    \vspace{-1em}
    \caption{Efficiency Scalability Analysis. (Left) Query-time latency $T_{latency}$ (ms) comparison between GRDR and dense retrieval across varying corpus sizes. (Right) Index storage requirements (MB) comparing GRDR's semantic IDs against dense frame-level and video-level features.}
    \label{fig:efficiency}
    \vspace{-5pt}
\end{figure}

\subsection{Efficiency Analysis (RQ2)}
\textbf{Inference Latency Analysis.} GRDR achieves $3\times$ to $15\times$ speedup over dense retrieval baselines in the Inductive Setting. Among GR methods, GRDR also achieves the lowest latency. TIGER \cite{GR_RecSys_TIGER} and AVG \cite{GR_IR_AVG} require a codebook configuration (4 layers, size 256) to fully quantize videos into diverse entries. Moreover, their implementations use a single shared vocabulary across all layers rather than independent codebooks. This results in vocabulary redundancy that increases decoding overhead. T2VIndexer \cite{GT_TVR_T2VIndexer} builds a compact codebook through k-means tree clustering, yet introduces a Prefix-aware decoder adapter that adds inference overhead. Furthermore, GRDR's multi-view tokenization provides multiple retrieval paths per video, enabling equivalent recall coverage with fewer beam candidates. EERCF \cite{TVR_EERCF} incurs the highest latency on inductive setting due to its cross-modal interaction reranking, which requires on-the-fly processing of frame-level and patch-level features. We observe that GRDR exhibits a slight latency increase from 136ms to 184ms on MSR-VTT Full-Corpus Setting. This stems from code collisions between training and test sets, which generate additional candidates requiring deduplication. Despite this minor increase, the efficiency gap over dense retrieval widens dramatically at scale. As shown in the left of Figure \ref{fig:efficiency}, dense retrieval latency grows linearly with corpus size while GRDR remains constant, achieving up to $300\times$ speedup on LSMDC \cite{LSMDC} with 0.1 million videos. 

\noindent \textbf{Index Storage Analysis.} As illustrated in Figure \ref{fig:efficiency} right, GRDR achieves substantial storage reduction compared to representative dense retrieval baselines. Specifically, GRDR requires $42\times$ less storage than video-level embeddings used by CLIP4Clip \cite{TVR_CLIP4clip} and $500\times$ less than frame-level embeddings used by X-Pool \cite{TVR_Xpool}. This reduction stems from replacing high-dimensional continuous embeddings with compact discrete semantic IDs. To simulate a practical large-scale scenario, we increase the GRDR codebook layers to 6 across 4 semantic views. In contrast, dense methods require storing 512-dimensional float vectors for each frame or video. Projecting to industry-scale deployment, a 1M video corpus requires only 46MB to store semantic IDs, enabling efficient retrieval on resource-constrained environments where dense embedding storage becomes prohibitive.

\begin{table}[!t]
\definecolor{lightyellow}{RGB}{255, 249, 192}
\definecolor{lightpurple}{RGB}{248, 231, 242}
\definecolor{deltared}{RGB}{255,150,150}
\definecolor{deltagreen}{RGB}{0, 150, 0}
\newcommand{\res}[2]{#1 \footnotesize{\textcolor{deltared}{(#2)}}}
\newcommand{\respos}[2]{#1 \footnotesize{\textcolor{deltagreen}{(+#2)}}}
\newcommand{\resneutral}[1]{#1 \footnotesize{\textcolor{gray}{(0.0)}}}
\caption{Ablation study of key components in GRDR. We evaluate the impact of Multi-view Tokenizer, Unified Co-training, Hierarchical Consistency Loss ($\mathcal{L}_{HC}$) and Cross-Modal Alignment Loss ($\mathcal{L}_{CL}$) on MSRVTT \cite{MSRVTT} and ACTNET \cite{ACTNET}.}
\centering
\small
\setlength{\aboverulesep}{0pt}
\setlength{\belowrulesep}{0pt}
\resizebox{0.475\textwidth}{!}{%
\begin{tabular}{l|ccc|ccc}
\toprule
\multicolumn{1}{l|}{\multirow{2}{*}{\textbf{Model}}} & \multicolumn{3}{c|}{\textbf{MSR-VTT \cite{MSRVTT}}} & \multicolumn{3}{c}{\textbf{ActivityNet \cite{ACTNET}}} \\
 & R@1 & R@5 & R@10 & R@1 & R@5 & R@10 \\
\midrule
\rowcolor{lightyellow}
\multicolumn{7}{l}{\textit{\textbf{Inductive Setting}}} \\
\rowcolor{lightpurple}
\textbf{GRDR (Ours)} & \textbf{46.0} & \textbf{70.1} & \textbf{78.0} & \textbf{33.6} & \textbf{63.6} & \textbf{76.5} \\
~~w/o Multi-view Tokenizer & \res{44.8}{-1.2} & 67.9 & 77.1 & \res{33.3}{-0.3} & 62.9 & 75.7 \\
~~w/o Unified Co-training & \res{44.5}{-1.5} & 66.8 & 75.8 & \res{32.6}{-1.0} & 62.1 & 74.5 \\
~~w/o $\mathcal{L}_{HC}$ & \res{43.3}{-2.7} & 65.8 & 73.9 & \res{32.7}{-0.9} & 61.2 & 73.9 \\
~~w/o $\mathcal{L}_{CL}$ & \res{38.9}{-7.1} & 58.5 & 65.7 & \res{24.1}{-9.5} & 44.8 & 53.9 \\
\midrule
\rowcolor{lightyellow}
\multicolumn{7}{l}{\textit{\textbf{Full-Corpus Setting}}} \\
\rowcolor{lightpurple}
\textbf{GRDR (Ours)} & \textbf{17.4} & \textbf{32.2} & \textbf{39.7} & \textbf{19.2} & \textbf{41.1} & \textbf{51.8} \\
~~w/o Multi-view Tokenizer & \res{15.0}{-2.4} & 29.0 & 35.6 & \res{18.5}{-0.7} & 39.1 & 50.0 \\
~~w/o Unified Co-training & \res{15.3}{-2.1} & 28.3 & 34.6 & \res{18.7}{-0.5} & 39.3 & 49.9 \\
~~w/o $\mathcal{L}_{HC}$ & \res{14.9}{-2.5} & 28.3 & 34.7 & \res{17.9}{-1.3} & 38.0 & 48.0 \\
~~w/o $\mathcal{L}_{CL}$ & \res{7.2}{-10.2} & 15.5 & 18.8 & \res{12.8}{-6.4} & 26.3 & 33.2 \\
\bottomrule
\end{tabular}%
}
\label{tab:ablation}
\vspace{-10pt}
\end{table}

\subsection{Effectiveness of Components (RQ3)}

To validate our architectural contributions, we conduct ablation studies and qualitative analysis to examine how each component affects retrieval performance.

\noindent \textbf{Effectiveness of Individual Components.} We systematically remove individual modules and present results in Table \ref{tab:ablation}. All components contribute meaningfully, as removing any single one leads to consistent performance degradation across both datasets and settings. \textbf{Multi-View Video Tokenizer} enables each video to be encoded into multiple semantic IDs. Replacing it with a standard single-view video tokenizer causes performance drops, with larger degradation under Full-Corpus setting. This indicates that multi-view encoding becomes increasingly important at scale, where diverse semantic access paths help distinguish similar videos. \textbf{Unified Co-Training} couples the tokenizer and retriever through a shared codebook for end-to-end optimization. Replacing it with a two-stage training pipeline degrades performance, confirming that bidirectional gradient flow allows retrieval objectives to directly refine video quantization. During unified co-training, \textbf{Hierarchical Consistency Loss ($\mathcal{L}_{HC}$)} prevents code drift during progressive training. Removal causes degradation on both settings as earlier codebook layers become unstable when later layers train. \textbf{Cross-Modal Alignment Loss ($\mathcal{L}_{CL}$)} bridges both components by aligning video latent features $f_v$ with decoder output features $h_m$ in a shared embedding space. Removing this loss causes the most severe degradation. Without alignment, codebook collision rates increase substantially, amplifying semantic ambiguity. This confirms that without explicit cross-modal supervision, the tokenizer produces codes optimized for visual reconstruction rather than textual relevance.

\begin{figure}
    \centering
    \includegraphics[width=1\linewidth]{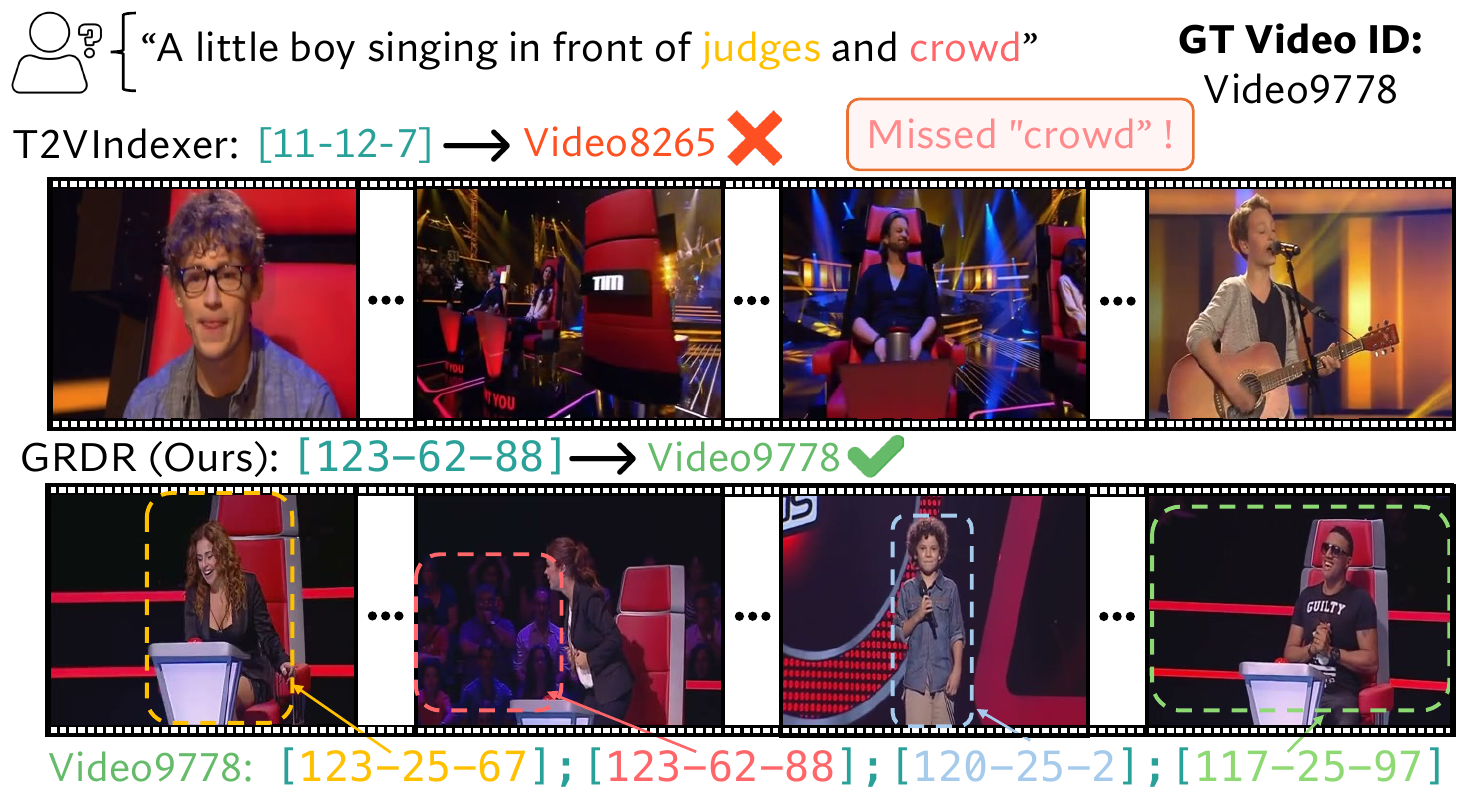}
    \caption{Case study of retrieval results. T2VIndexer \cite{GT_TVR_T2VIndexer} (Top) retrieves a video but only matching partial the subject. GRDR (Bottom) retrieves the ground-truth video containing the specific details from the query.}
    \label{fig:case_study}
    \vspace{-0.5em}
\end{figure}

\begin{figure}
    \centering
    \includegraphics[width=1\linewidth]{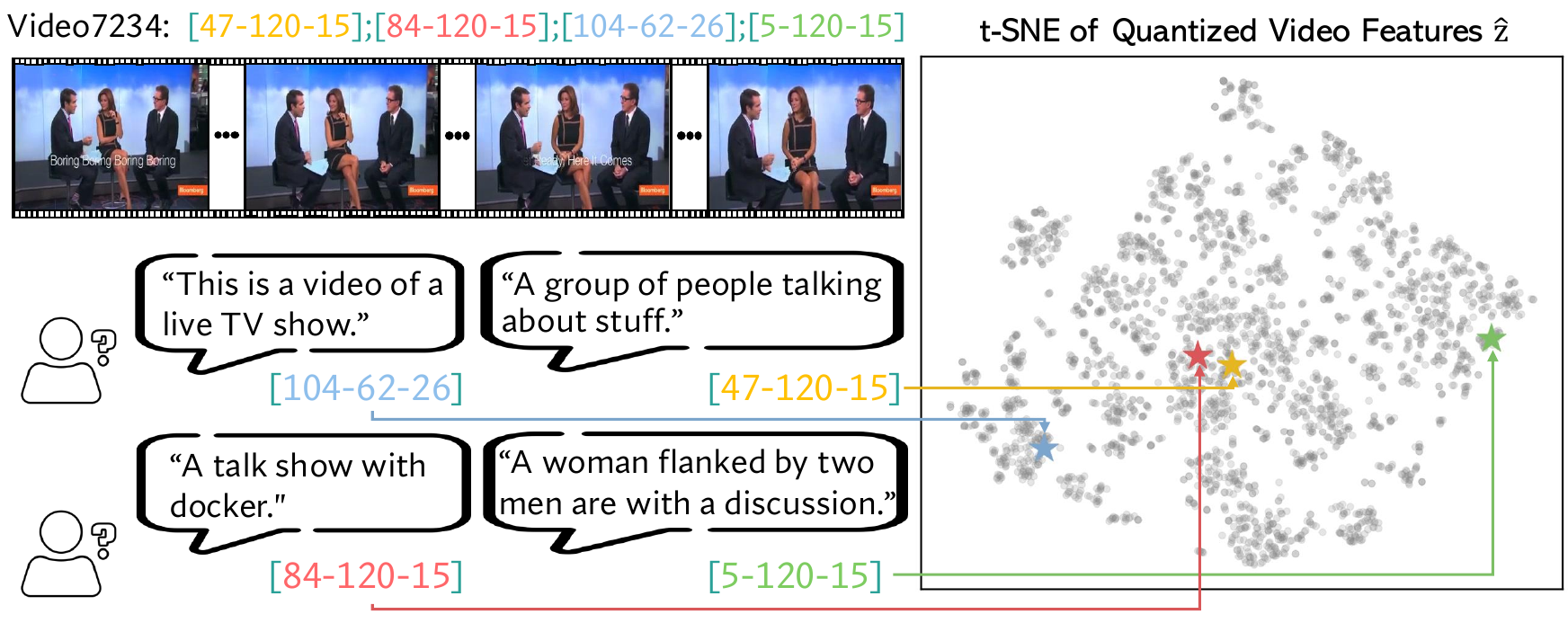}
    \caption{Visualization of our Multi-View Video Tokenizer. This t-SNE plot visualizes the quantized video features $\hat{\mathbf{z}}$ of all semantic IDs generated from the MSR-VTT \cite{MSRVTT} test set.}
    \label{fig:tsne}
    \vspace{-0.5em}
\end{figure}

\noindent \textbf{Case Study.} Figure \ref{fig:case_study} presents a case on a complex query requiring multiple semantic elements. T2VIndexer \cite{GT_TVR_T2VIndexer} employs one-to-one mapping, retrieving a video matching only the main subject while missing contextual details. In contrast, GRDR encodes the ground-truth video into four distinct semantic IDs, each capturing a different facet: `judges', `crowd', `singing boy', and `show context'. When the query specifically mentions 'crowd', GRDR successfully retrieves the video via the corresponding semantic view. This demonstrates that multi-view encoding resolves semantic ambiguity, enabling accurate retrieval even for highly specific queries.

\noindent \textbf{Qualitative Analysis of Multi-View Semantic IDs.} To further validate that our multi-view tokenizer captures genuinely distinct semantics, Figure \ref{fig:tsne} visualizes the quantized video features $\hat{z}$ via t-SNE. Four diverse queries successfully retrieve Video7234 through these different semantic IDs, which demonstrates GRDR's ability to recall the same video from multiple semantic perspectives. 
This spatial distribution confirms that query-guided encoding effectively diversifies semantic representations.

\begin{figure}[!t]
    \centering
    \includegraphics[width=1\linewidth]{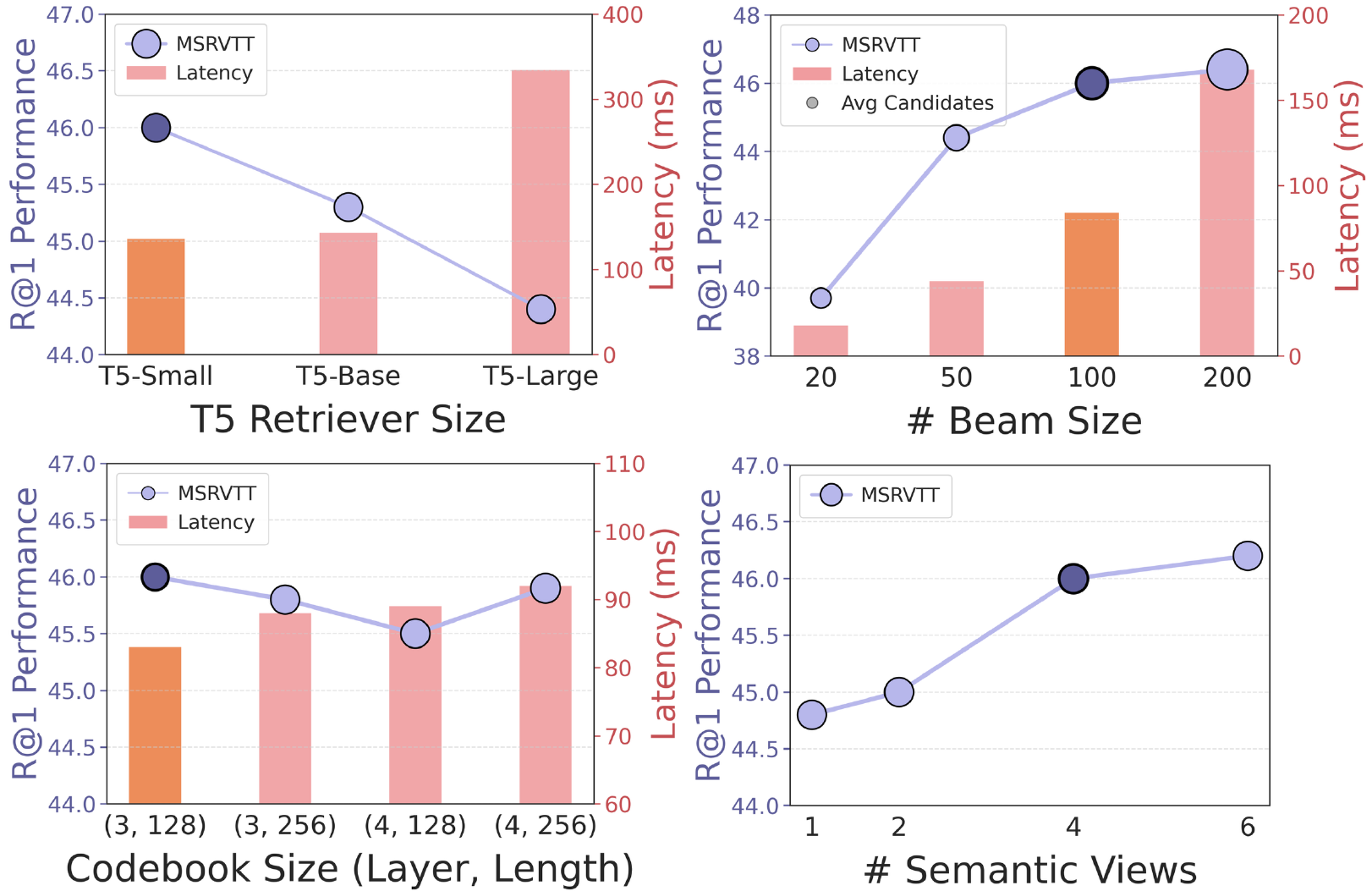}
    \caption{Hyper-parameter sensitivity on MSRVTT \cite{MSRVTT} dataset under inductive setting.}
    \label{fig:hyper}
    \vspace{-0.5em}
\end{figure}

\subsection{Hyper-parameter Study (RQ4)}
We conducted a series of experiments to investigate the impact of the four primary hyper-parameters on the trade-off between the model's effectiveness and efficiency.

\noindent \textbf{Impact of T5 Retriever backbone Size.}
We evaluate three T5 variants \cite{T5}: T5-Small (60M), T5-Base (220M), and T5-Large (800M). As shown in Figure~\ref{fig:hyper} (top-left), increasing model size yields no performance gains while inference latency triples for T5-Large. This indicates T5-Small sufficiently captures query semantics. Our Unified Co-training drives this efficiency by aligning components in a shared semantic space, simplifying semantic ID prediction without requiring larger model capacity.

\noindent \textbf{Number of Beams Size.}
We vary beam size from 20 to 200, with circle size indicating average candidates generated. Performance improves rapidly initially but plateaus after 100 beams, while latency continues growing. We select beam size 100 as the optimal balance between recall coverage and reranking cost.

\noindent \textbf{Impact of Codebook Size.}
We analyze codebook dimensions (layers and length) in Figure~\ref{fig:hyper} (bottom-left). Increasing dimensions yields no significant gains, indicating that Multi-View semantic IDs and Unified Co-training effectively capture complex video semantics even with highly compressed code space. Since latency increases with codebook size, we select configuration (3, 128) to maximize both efficiency and effectiveness.

\noindent \textbf{Number of Semantic Views.}
We vary $N_v$ from 1 to 6. Performance improves initially but saturates beyond 4 views. Too few semantic IDs cause semantic ambiguity by failing to capture diverse user intents, while excessive views result in redundancy and increased semantic ID collisions. We identify $N_v=4$ as optimal configuration.

\section{Conclusion}
We present GRDR, a Generative Recall and Dense Reranking framework that addresses two fundamental challenges limiting generative retrieval for TVR: (i) semantic ambiguity, where polysemous videos are forced into single semantic IDs, and (ii) cross-modal misalignment, where semantic IDs lack supervision from text queries. To resolve these challenges, GRDR introduces a Multi-View Video Tokenizer that assigns multiple semantic IDs per video via query-guided contrastive learning, and a Unified Co-Training scheme that couples tokenizer and retriever through a shared codebook for end-to-end optimization. Experiments on four benchmarks demonstrate that GRDR achieves competitive accuracy with dense retrievers while reducing storage by an order of magnitude and accelerating inference by up to $300\times$. Our established GRDR provides an effective generative retrieval recall stage that bridges the gap between scalability and accuracy, paving the way for future research in efficient text-to-video retrieval.



\newpage
\balance
\bibliographystyle{ACM-Reference-Format}
\bibliography{reference}


\begin{thebibliography}{56}


\ifx \showCODEN    \undefined \def \showCODEN     #1{\unskip}     \fi
\ifx \showISBNx    \undefined \def \showISBNx     #1{\unskip}     \fi
\ifx \showISBNxiii \undefined \def \showISBNxiii  #1{\unskip}     \fi
\ifx \showISSN     \undefined \def \showISSN      #1{\unskip}     \fi
\ifx \showLCCN     \undefined \def \showLCCN      #1{\unskip}     \fi
\ifx \shownote     \undefined \def \shownote      #1{#1}          \fi
\ifx \showarticletitle \undefined \def \showarticletitle #1{#1}   \fi
\ifx \showURL      \undefined \def \showURL       {\relax}        \fi
\providecommand\bibfield[2]{#2}
\providecommand\bibinfo[2]{#2}
\providecommand\natexlab[1]{#1}
\providecommand\showeprint[2][]{arXiv:#2}

\bibitem[noa({[n.\,d.]})]%
        {noauthor_gpt-5_nodate}
 \bibinfo{year}{[n.\,d.]}\natexlab{}.
\newblock \bibinfo{title}{{GPT}-5 mini {Model} {\textbar} {OpenAI} {API}}.
\newblock
\urldef\tempurl%
\url{https://platform.openai.com}
\showURL{%
\tempurl}


\bibitem[Bai et~al\mbox{.}(2025)]%
        {VLM3}
\bibfield{author}{\bibinfo{person}{Shuai Bai}, \bibinfo{person}{Keqin Chen}, \bibinfo{person}{Xuejing Liu}, \bibinfo{person}{Jialin Wang}, \bibinfo{person}{Wenbin Ge}, \bibinfo{person}{Sibo Song}, \bibinfo{person}{Kai Dang}, \bibinfo{person}{Peng Wang}, \bibinfo{person}{Shijie Wang}, \bibinfo{person}{Jun Tang}, \bibinfo{person}{Humen Zhong}, \bibinfo{person}{Yuanzhi Zhu}, \bibinfo{person}{Ming{-}Hsuan Yang}, \bibinfo{person}{Zhaohai Li}, \bibinfo{person}{Jianqiang Wan}, \bibinfo{person}{Pengfei Wang}, \bibinfo{person}{Wei Ding}, \bibinfo{person}{Zheren Fu}, \bibinfo{person}{Yiheng Xu}, \bibinfo{person}{Jiabo Ye}, \bibinfo{person}{Xi Zhang}, \bibinfo{person}{Tianbao Xie}, \bibinfo{person}{Zesen Cheng}, \bibinfo{person}{Hang Zhang}, \bibinfo{person}{Zhibo Yang}, \bibinfo{person}{Haiyang Xu}, {and} \bibinfo{person}{Junyang Lin}.} \bibinfo{year}{2025}\natexlab{}.
\newblock \showarticletitle{Qwen2.5-VL Technical Report}.
\newblock \bibinfo{journal}{\emph{CoRR}}  \bibinfo{volume}{abs/2502.13923} (\bibinfo{year}{2025}).
\newblock
\showeprint[arXiv]{2502.13923}
\href{https://doi.org/10.48550/ARXIV.2502.13923}{doi:\nolinkurl{10.48550/ARXIV.2502.13923}}


\bibitem[Chen et~al\mbox{.}(2023)]%
        {VLM2}
\bibfield{author}{\bibinfo{person}{Zhe Chen}, \bibinfo{person}{Jiannan Wu}, \bibinfo{person}{Wenhai Wang}, \bibinfo{person}{Weijie Su}, \bibinfo{person}{Guo Chen}, \bibinfo{person}{Sen Xing}, \bibinfo{person}{Muyan Zhong}, \bibinfo{person}{Qinglong Zhang}, \bibinfo{person}{Xizhou Zhu}, \bibinfo{person}{Lewei Lu}, \bibinfo{person}{Bin Li}, \bibinfo{person}{Ping Luo}, \bibinfo{person}{Tong Lu}, \bibinfo{person}{Yu Qiao}, {and} \bibinfo{person}{Jifeng Dai}.} \bibinfo{year}{2023}\natexlab{}.
\newblock \showarticletitle{InternVL: Scaling up Vision Foundation Models and Aligning for Generic Visual-Linguistic Tasks}.
\newblock \bibinfo{journal}{\emph{CoRR}}  \bibinfo{volume}{abs/2312.14238} (\bibinfo{year}{2023}).
\newblock
\showeprint[arXiv]{2312.14238}
\href{https://doi.org/10.48550/ARXIV.2312.14238}{doi:\nolinkurl{10.48550/ARXIV.2312.14238}}


\bibitem[Chen et~al\mbox{.}(2025)]%
        {CV1}
\bibfield{author}{\bibinfo{person}{Zhi Chen}, \bibinfo{person}{Zecheng Zhao}, \bibinfo{person}{Jingcai Guo}, \bibinfo{person}{Jingjing Li}, {and} \bibinfo{person}{Zi Huang}.} \bibinfo{year}{2025}\natexlab{}.
\newblock \showarticletitle{{SVIP:} Semantically Contextualized Visual Patches for Zero-Shot Learning}.
\newblock \bibinfo{journal}{\emph{CoRR}}  \bibinfo{volume}{abs/2503.10252} (\bibinfo{year}{2025}).
\newblock


\bibitem[Covington et~al\mbox{.}(2016)]%
        {RecallRerank1}
\bibfield{author}{\bibinfo{person}{Paul Covington}, \bibinfo{person}{Jay Adams}, {and} \bibinfo{person}{Emre Sargin}.} \bibinfo{year}{2016}\natexlab{}.
\newblock \showarticletitle{Deep Neural Networks for YouTube Recommendations}. In \bibinfo{booktitle}{\emph{RecSys}}. \bibinfo{publisher}{{ACM}}, \bibinfo{pages}{191--198}.
\newblock


\bibitem[Datar et~al\mbox{.}(2004)]%
        {LSH}
\bibfield{author}{\bibinfo{person}{Mayur Datar}, \bibinfo{person}{Nicole Immorlica}, \bibinfo{person}{Piotr Indyk}, {and} \bibinfo{person}{Vahab~S. Mirrokni}.} \bibinfo{year}{2004}\natexlab{}.
\newblock \showarticletitle{Locality-sensitive hashing scheme based on p-stable distributions}. In \bibinfo{booktitle}{\emph{{SCG}}}. \bibinfo{publisher}{{ACM}}, \bibinfo{pages}{253--262}.
\newblock


\bibitem[Deng et~al\mbox{.}(2023)]%
        {TVR_Prompt_Switch}
\bibfield{author}{\bibinfo{person}{Chaorui Deng}, \bibinfo{person}{Qi Chen}, \bibinfo{person}{Pengda Qin}, \bibinfo{person}{Da Chen}, {and} \bibinfo{person}{Qi Wu}.} \bibinfo{year}{2023}\natexlab{}.
\newblock \showarticletitle{Prompt Switch: Efficient {CLIP} Adaptation for Text-Video Retrieval}. In \bibinfo{booktitle}{\emph{{IEEE/CVF} International Conference on Computer Vision, {ICCV} 2023, Paris, France, October 1-6, 2023}}. \bibinfo{publisher}{{IEEE}}, \bibinfo{pages}{15602--15612}.
\newblock
\href{https://doi.org/10.1109/ICCV51070.2023.01434}{doi:\nolinkurl{10.1109/ICCV51070.2023.01434}}


\bibitem[Ding et~al\mbox{.}(2024)]%
        {GR_RecSys2}
\bibfield{author}{\bibinfo{person}{Yijie Ding}, \bibinfo{person}{Yupeng Hou}, \bibinfo{person}{Jiacheng Li}, {and} \bibinfo{person}{Julian~J. McAuley}.} \bibinfo{year}{2024}\natexlab{}.
\newblock \showarticletitle{Inductive Generative Recommendation via Retrieval-based Speculation}.
\newblock \bibinfo{journal}{\emph{CoRR}}  \bibinfo{volume}{abs/2410.02939} (\bibinfo{year}{2024}).
\newblock
\showeprint[arXiv]{2410.02939}
\href{https://doi.org/10.48550/ARXIV.2410.02939}{doi:\nolinkurl{10.48550/ARXIV.2410.02939}}


\bibitem[Fu et~al\mbox{.}(2025)]%
        {CLIP3}
\bibfield{author}{\bibinfo{person}{Yuxia Fu}, \bibinfo{person}{Zhizhen Zhang}, \bibinfo{person}{Yuqi Zhang}, \bibinfo{person}{Zijian Wang}, \bibinfo{person}{Zi Huang}, {and} \bibinfo{person}{Yadan Luo}.} \bibinfo{year}{2025}\natexlab{}.
\newblock \showarticletitle{MergeVLA: Cross-Skill Model Merging Toward a Generalist Vision-Language-Action Agent}.
\newblock \bibinfo{journal}{\emph{CoRR}}  \bibinfo{volume}{abs/2511.18810} (\bibinfo{year}{2025}).
\newblock


\bibitem[Gorti et~al\mbox{.}(2022)]%
        {TVR_Xpool}
\bibfield{author}{\bibinfo{person}{Satya~Krishna Gorti}, \bibinfo{person}{No{\"{e}}l Vouitsis}, \bibinfo{person}{Junwei Ma}, \bibinfo{person}{Keyvan Golestan}, \bibinfo{person}{Maksims Volkovs}, \bibinfo{person}{Animesh Garg}, {and} \bibinfo{person}{Guangwei Yu}.} \bibinfo{year}{2022}\natexlab{}.
\newblock \showarticletitle{X-Pool: Cross-Modal Language-Video Attention for Text-Video Retrieval}. In \bibinfo{booktitle}{\emph{{IEEE/CVF} Conference on Computer Vision and Pattern Recognition, {CVPR} 2022, New Orleans, LA, USA, June 18-24, 2022}}. \bibinfo{publisher}{{IEEE}}, \bibinfo{pages}{4996--5005}.
\newblock
\href{https://doi.org/10.1109/CVPR52688.2022.00495}{doi:\nolinkurl{10.1109/CVPR52688.2022.00495}}


\bibitem[Heilbron et~al\mbox{.}(2015)]%
        {ACTNET}
\bibfield{author}{\bibinfo{person}{Fabian~Caba Heilbron}, \bibinfo{person}{Victor Escorcia}, \bibinfo{person}{Bernard Ghanem}, {and} \bibinfo{person}{Juan~Carlos Niebles}.} \bibinfo{year}{2015}\natexlab{}.
\newblock \showarticletitle{ActivityNet: {A} large-scale video benchmark for human activity understanding}. In \bibinfo{booktitle}{\emph{{IEEE} Conference on Computer Vision and Pattern Recognition, {CVPR} 2015, Boston, MA, USA, June 7-12, 2015}}. \bibinfo{publisher}{{IEEE} Computer Society}, \bibinfo{pages}{961--970}.
\newblock
\href{https://doi.org/10.1109/CVPR.2015.7298698}{doi:\nolinkurl{10.1109/CVPR.2015.7298698}}


\bibitem[Hendricks et~al\mbox{.}(2017)]%
        {DiDeMo}
\bibfield{author}{\bibinfo{person}{Lisa~Anne Hendricks}, \bibinfo{person}{Oliver Wang}, \bibinfo{person}{Eli Shechtman}, \bibinfo{person}{Josef Sivic}, \bibinfo{person}{Trevor Darrell}, {and} \bibinfo{person}{Bryan~C. Russell}.} \bibinfo{year}{2017}\natexlab{}.
\newblock \showarticletitle{Localizing Moments in Video with Natural Language}. In \bibinfo{booktitle}{\emph{{IEEE} International Conference on Computer Vision, {ICCV} 2017, Venice, Italy, October 22-29, 2017}}. \bibinfo{publisher}{{IEEE} Computer Society}, \bibinfo{pages}{5804--5813}.
\newblock
\href{https://doi.org/10.1109/ICCV.2017.618}{doi:\nolinkurl{10.1109/ICCV.2017.618}}


\bibitem[Hou et~al\mbox{.}(2025)]%
        {GR_RecSys}
\bibfield{author}{\bibinfo{person}{Yupeng Hou}, \bibinfo{person}{Jianmo Ni}, \bibinfo{person}{Zhankui He}, \bibinfo{person}{Noveen Sachdeva}, \bibinfo{person}{Wang{-}Cheng Kang}, \bibinfo{person}{Ed~H. Chi}, \bibinfo{person}{Julian~J. McAuley}, {and} \bibinfo{person}{Derek~Zhiyuan Cheng}.} \bibinfo{year}{2025}\natexlab{}.
\newblock \showarticletitle{ActionPiece: Contextually Tokenizing Action Sequences for Generative Recommendation}. In \bibinfo{booktitle}{\emph{Forty-second International Conference on Machine Learning, {ICML} 2025, Vancouver, BC, Canada, July 13-19, 2025}}. \bibinfo{publisher}{OpenReview.net}.
\newblock
\urldef\tempurl%
\url{https://openreview.net/forum?id=h2oNQOzbc5}
\showURL{%
\tempurl}


\bibitem[Huang et~al\mbox{.}(2020)]%
        {Recall_Rerank2}
\bibfield{author}{\bibinfo{person}{Jui{-}Ting Huang}, \bibinfo{person}{Ashish Sharma}, \bibinfo{person}{Shuying Sun}, \bibinfo{person}{Li Xia}, \bibinfo{person}{David Zhang}, \bibinfo{person}{Philip Pronin}, \bibinfo{person}{Janani Padmanabhan}, \bibinfo{person}{Giuseppe Ottaviano}, {and} \bibinfo{person}{Linjun Yang}.} \bibinfo{year}{2020}\natexlab{}.
\newblock \showarticletitle{Embedding-based Retrieval in Facebook Search}. In \bibinfo{booktitle}{\emph{{KDD}}}. \bibinfo{publisher}{{ACM}}, \bibinfo{pages}{2553--2561}.
\newblock


\bibitem[Jin et~al\mbox{.}(2024a)]%
        {GR_DR}
\bibfield{author}{\bibinfo{person}{Bowen Jin}, \bibinfo{person}{Hansi Zeng}, \bibinfo{person}{Guoyin Wang}, \bibinfo{person}{Xiusi Chen}, \bibinfo{person}{Tianxin Wei}, \bibinfo{person}{Ruirui Li}, \bibinfo{person}{Zhengyang Wang}, \bibinfo{person}{Zheng Li}, \bibinfo{person}{Yang Li}, \bibinfo{person}{Hanqing Lu}, \bibinfo{person}{Suhang Wang}, \bibinfo{person}{Jiawei Han}, {and} \bibinfo{person}{Xianfeng Tang}.} \bibinfo{year}{2024}\natexlab{a}.
\newblock \showarticletitle{Language Models as Semantic Indexers}. In \bibinfo{booktitle}{\emph{Forty-first International Conference on Machine Learning, {ICML} 2024, Vienna, Austria, July 21-27, 2024}}. \bibinfo{publisher}{OpenReview.net}.
\newblock
\urldef\tempurl%
\url{https://openreview.net/forum?id=sYeioWoF9u}
\showURL{%
\tempurl}


\bibitem[Jin et~al\mbox{.}(2023)]%
        {TVR_DiffusionRet}
\bibfield{author}{\bibinfo{person}{Peng Jin}, \bibinfo{person}{Hao Li}, \bibinfo{person}{Zesen Cheng}, \bibinfo{person}{Kehan Li}, \bibinfo{person}{Xiangyang Ji}, \bibinfo{person}{Chang Liu}, \bibinfo{person}{Li Yuan}, {and} \bibinfo{person}{Jie Chen}.} \bibinfo{year}{2023}\natexlab{}.
\newblock \showarticletitle{DiffusionRet: Generative Text-Video Retrieval with Diffusion Model}. In \bibinfo{booktitle}{\emph{{IEEE/CVF} International Conference on Computer Vision, {ICCV} 2023, Paris, France, October 1-6, 2023}}. \bibinfo{publisher}{{IEEE}}, \bibinfo{pages}{2470--2481}.
\newblock
\href{https://doi.org/10.1109/ICCV51070.2023.00234}{doi:\nolinkurl{10.1109/ICCV51070.2023.00234}}


\bibitem[Jin et~al\mbox{.}(2024b)]%
        {TVR_MV-Adapter}
\bibfield{author}{\bibinfo{person}{Xiaojie Jin}, \bibinfo{person}{Bowen Zhang}, \bibinfo{person}{Weibo Gong}, \bibinfo{person}{Kai Xu}, \bibinfo{person}{Xueqing Deng}, \bibinfo{person}{Peng Wang}, \bibinfo{person}{Zhao Zhang}, \bibinfo{person}{Xiaohui Shen}, {and} \bibinfo{person}{Jiashi Feng}.} \bibinfo{year}{2024}\natexlab{b}.
\newblock \showarticletitle{MV-Adapter: Multimodal Video Transfer Learning for Video Text Retrieval}. In \bibinfo{booktitle}{\emph{{IEEE/CVF} Conference on Computer Vision and Pattern Recognition, {CVPR} 2024, Seattle, WA, USA, June 16-22, 2024}}. \bibinfo{publisher}{{IEEE}}, \bibinfo{pages}{27134--27143}.
\newblock
\href{https://doi.org/10.1109/CVPR52733.2024.02563}{doi:\nolinkurl{10.1109/CVPR52733.2024.02563}}


\bibitem[Kim et~al\mbox{.}(2025)]%
        {GR_MM}
\bibfield{author}{\bibinfo{person}{Sungyeon Kim}, \bibinfo{person}{Xinliang Zhu}, \bibinfo{person}{Xiaofan Lin}, \bibinfo{person}{Muhammet Bastan}, \bibinfo{person}{Douglas Gray}, {and} \bibinfo{person}{Suha Kwak}.} \bibinfo{year}{2025}\natexlab{}.
\newblock \showarticletitle{{GENIUS:} {A} Generative Framework for Universal Multimodal Search}. In \bibinfo{booktitle}{\emph{{IEEE/CVF} Conference on Computer Vision and Pattern Recognition, {CVPR} 2025, Nashville, TN, USA, June 11-15, 2025}}. \bibinfo{publisher}{Computer Vision Foundation / {IEEE}}, \bibinfo{pages}{19659--19669}.
\newblock
\href{https://doi.org/10.1109/CVPR52734.2025.01831}{doi:\nolinkurl{10.1109/CVPR52734.2025.01831}}


\bibitem[Ko et~al\mbox{.}(2025)]%
        {BLIM}
\bibfield{author}{\bibinfo{person}{Dohwan Ko}, \bibinfo{person}{Ji~Soo Lee}, \bibinfo{person}{Minhyuk Choi}, \bibinfo{person}{Zihang Meng}, {and} \bibinfo{person}{Hyunwoo~J. Kim}.} \bibinfo{year}{2025}\natexlab{}.
\newblock \showarticletitle{Bidirectional Likelihood Estimation with Multi-Modal Large Language Models for Text-Video Retrieval}.
\newblock \bibinfo{journal}{\emph{CoRR}}  \bibinfo{volume}{abs/2507.23284} (\bibinfo{year}{2025}).
\newblock
\showeprint[arXiv]{2507.23284}
\href{https://doi.org/10.48550/ARXIV.2507.23284}{doi:\nolinkurl{10.48550/ARXIV.2507.23284}}


\bibitem[Lee et~al\mbox{.}(2022)]%
        {RQ}
\bibfield{author}{\bibinfo{person}{Doyup Lee}, \bibinfo{person}{Chiheon Kim}, \bibinfo{person}{Saehoon Kim}, \bibinfo{person}{Minsu Cho}, {and} \bibinfo{person}{Wook{-}Shin Han}.} \bibinfo{year}{2022}\natexlab{}.
\newblock \showarticletitle{Autoregressive Image Generation using Residual Quantization}. In \bibinfo{booktitle}{\emph{{IEEE/CVF} Conference on Computer Vision and Pattern Recognition, {CVPR} 2022, New Orleans, LA, USA, June 18-24, 2022}}. \bibinfo{publisher}{{IEEE}}, \bibinfo{pages}{11513--11522}.
\newblock
\href{https://doi.org/10.1109/CVPR52688.2022.01123}{doi:\nolinkurl{10.1109/CVPR52688.2022.01123}}


\bibitem[Lei et~al\mbox{.}(2021)]%
        {TVR_CLIPBert}
\bibfield{author}{\bibinfo{person}{Jie Lei}, \bibinfo{person}{Linjie Li}, \bibinfo{person}{Luowei Zhou}, \bibinfo{person}{Zhe Gan}, \bibinfo{person}{Tamara~L. Berg}, \bibinfo{person}{Mohit Bansal}, {and} \bibinfo{person}{Jingjing Liu}.} \bibinfo{year}{2021}\natexlab{}.
\newblock \showarticletitle{Less Is More: ClipBERT for Video-and-Language Learning via Sparse Sampling}. In \bibinfo{booktitle}{\emph{{IEEE} Conference on Computer Vision and Pattern Recognition, {CVPR} 2021, virtual, June 19-25, 2021}}. \bibinfo{publisher}{Computer Vision Foundation / {IEEE}}, \bibinfo{pages}{7331--7341}.
\newblock
\href{https://doi.org/10.1109/CVPR46437.2021.00725}{doi:\nolinkurl{10.1109/CVPR46437.2021.00725}}


\bibitem[Li et~al\mbox{.}(2025b)]%
        {VLM1}
\bibfield{author}{\bibinfo{person}{Bo Li}, \bibinfo{person}{Yuanhan Zhang}, \bibinfo{person}{Dong Guo}, \bibinfo{person}{Renrui Zhang}, \bibinfo{person}{Feng Li}, \bibinfo{person}{Hao Zhang}, \bibinfo{person}{Kaichen Zhang}, \bibinfo{person}{Peiyuan Zhang}, \bibinfo{person}{Yanwei Li}, \bibinfo{person}{Ziwei Liu}, {and} \bibinfo{person}{Chunyuan Li}.} \bibinfo{year}{2025}\natexlab{b}.
\newblock \showarticletitle{LLaVA-OneVision: Easy Visual Task Transfer}.
\newblock \bibinfo{journal}{\emph{Trans. Mach. Learn. Res.}}  \bibinfo{volume}{2025} (\bibinfo{year}{2025}).
\newblock
\urldef\tempurl%
\url{https://openreview.net/forum?id=zKv8qULV6n}
\showURL{%
\tempurl}


\bibitem[Li et~al\mbox{.}(2022)]%
        {BLIP}
\bibfield{author}{\bibinfo{person}{Junnan Li}, \bibinfo{person}{Dongxu Li}, \bibinfo{person}{Caiming Xiong}, {and} \bibinfo{person}{Steven C.~H. Hoi}.} \bibinfo{year}{2022}\natexlab{}.
\newblock \showarticletitle{{BLIP:} Bootstrapping Language-Image Pre-training for Unified Vision-Language Understanding and Generation}. In \bibinfo{booktitle}{\emph{International Conference on Machine Learning, {ICML} 2022, 17-23 July 2022, Baltimore, Maryland, {USA}}} \emph{(\bibinfo{series}{Proceedings of Machine Learning Research}, Vol.~\bibinfo{volume}{162})}, \bibfield{editor}{\bibinfo{person}{Kamalika Chaudhuri}, \bibinfo{person}{Stefanie Jegelka}, \bibinfo{person}{Le~Song}, \bibinfo{person}{Csaba Szepesv{\'{a}}ri}, \bibinfo{person}{Gang Niu}, {and} \bibinfo{person}{Sivan Sabato}} (Eds.). \bibinfo{publisher}{{PMLR}}, \bibinfo{pages}{12888--12900}.
\newblock
\urldef\tempurl%
\url{https://proceedings.mlr.press/v162/li22n.html}
\showURL{%
\tempurl}


\bibitem[Li et~al\mbox{.}(2023a)]%
        {TVR_ProST}
\bibfield{author}{\bibinfo{person}{Pandeng Li}, \bibinfo{person}{Chen{-}Wei Xie}, \bibinfo{person}{Liming Zhao}, \bibinfo{person}{Hongtao Xie}, \bibinfo{person}{Jiannan Ge}, \bibinfo{person}{Yun Zheng}, \bibinfo{person}{Deli Zhao}, {and} \bibinfo{person}{Yongdong Zhang}.} \bibinfo{year}{2023}\natexlab{a}.
\newblock \showarticletitle{Progressive Spatio-Temporal Prototype Matching for Text-Video Retrieval}. In \bibinfo{booktitle}{\emph{{IEEE/CVF} International Conference on Computer Vision, {ICCV} 2023, Paris, France, October 1-6, 2023}}. \bibinfo{publisher}{{IEEE}}, \bibinfo{pages}{4077--4087}.
\newblock
\href{https://doi.org/10.1109/ICCV51070.2023.00379}{doi:\nolinkurl{10.1109/ICCV51070.2023.00379}}


\bibitem[Li et~al\mbox{.}(2025a)]%
        {GR_IR_AVG}
\bibfield{author}{\bibinfo{person}{Yongqi Li}, \bibinfo{person}{Hongru Cai}, \bibinfo{person}{Wenjie Wang}, \bibinfo{person}{Leigang Qu}, \bibinfo{person}{Yinwei Wei}, \bibinfo{person}{Wenjie Li}, \bibinfo{person}{Liqiang Nie}, {and} \bibinfo{person}{Tat{-}Seng Chua}.} \bibinfo{year}{2025}\natexlab{a}.
\newblock \showarticletitle{Revolutionizing Text-to-Image Retrieval as Autoregressive Token-to-Voken Generation}. In \bibinfo{booktitle}{\emph{Proceedings of the 48th International {ACM} {SIGIR} Conference on Research and Development in Information Retrieval, {SIGIR} 2025, Padua, Italy, July 13-18, 2025}}, \bibfield{editor}{\bibinfo{person}{Nicola Ferro}, \bibinfo{person}{Maria Maistro}, \bibinfo{person}{Gabriella Pasi}, \bibinfo{person}{Omar Alonso}, \bibinfo{person}{Andrew Trotman}, {and} \bibinfo{person}{Suzan Verberne}} (Eds.). \bibinfo{publisher}{{ACM}}, \bibinfo{pages}{813--822}.
\newblock
\href{https://doi.org/10.1145/3726302.3730077}{doi:\nolinkurl{10.1145/3726302.3730077}}


\bibitem[Li et~al\mbox{.}(2024a)]%
        {GR_IR_GRACE}
\bibfield{author}{\bibinfo{person}{Yongqi Li}, \bibinfo{person}{Wenjie Wang}, \bibinfo{person}{Leigang Qu}, \bibinfo{person}{Liqiang Nie}, \bibinfo{person}{Wenjie Li}, {and} \bibinfo{person}{Tat{-}Seng Chua}.} \bibinfo{year}{2024}\natexlab{a}.
\newblock \showarticletitle{Generative Cross-Modal Retrieval: Memorizing Images in Multimodal Language Models for Retrieval and Beyond}. In \bibinfo{booktitle}{\emph{Proceedings of the 62nd Annual Meeting of the Association for Computational Linguistics (Volume 1: Long Papers), {ACL} 2024, Bangkok, Thailand, August 11-16, 2024}}, \bibfield{editor}{\bibinfo{person}{Lun{-}Wei Ku}, \bibinfo{person}{Andre Martins}, {and} \bibinfo{person}{Vivek Srikumar}} (Eds.). \bibinfo{publisher}{Association for Computational Linguistics}, \bibinfo{pages}{11851--11861}.
\newblock
\href{https://doi.org/10.18653/V1/2024.ACL-LONG.639}{doi:\nolinkurl{10.18653/V1/2024.ACL-LONG.639}}


\bibitem[Li et~al\mbox{.}(2023b)]%
        {GR_DR4}
\bibfield{author}{\bibinfo{person}{Yongqi Li}, \bibinfo{person}{Nan Yang}, \bibinfo{person}{Liang Wang}, \bibinfo{person}{Furu Wei}, {and} \bibinfo{person}{Wenjie Li}.} \bibinfo{year}{2023}\natexlab{b}.
\newblock \showarticletitle{Multiview Identifiers Enhanced Generative Retrieval}. In \bibinfo{booktitle}{\emph{{ACL} {(1)}}}. \bibinfo{publisher}{Association for Computational Linguistics}, \bibinfo{pages}{6636--6648}.
\newblock


\bibitem[Li et~al\mbox{.}(2024b)]%
        {GT_TVR_T2VIndexer}
\bibfield{author}{\bibinfo{person}{Yili Li}, \bibinfo{person}{Jing Yu}, \bibinfo{person}{Keke Gai}, \bibinfo{person}{Bang Liu}, \bibinfo{person}{Gang Xiong}, {and} \bibinfo{person}{Qi Wu}.} \bibinfo{year}{2024}\natexlab{b}.
\newblock \showarticletitle{T2VIndexer: {A} Generative Video Indexer for Efficient Text-Video Retrieval}. In \bibinfo{booktitle}{\emph{Proceedings of the 32nd {ACM} International Conference on Multimedia, {MM} 2024, Melbourne, VIC, Australia, 28 October 2024 - 1 November 2024}}, \bibfield{editor}{\bibinfo{person}{Jianfei Cai}, \bibinfo{person}{Mohan~S. Kankanhalli}, \bibinfo{person}{Balakrishnan Prabhakaran}, \bibinfo{person}{Susanne Boll}, \bibinfo{person}{Ramanathan Subramanian}, \bibinfo{person}{Liang Zheng}, \bibinfo{person}{Vivek~K. Singh}, \bibinfo{person}{Pablo C{\'{e}}sar}, \bibinfo{person}{Lexing Xie}, {and} \bibinfo{person}{Dong Xu}} (Eds.). \bibinfo{publisher}{{ACM}}, \bibinfo{pages}{3955--3963}.
\newblock
\href{https://doi.org/10.1145/3664647.3680673}{doi:\nolinkurl{10.1145/3664647.3680673}}


\bibitem[Liu et~al\mbox{.}(2025)]%
        {GR_RecSys_ETEGRec}
\bibfield{author}{\bibinfo{person}{Enze Liu}, \bibinfo{person}{Bowen Zheng}, \bibinfo{person}{Cheng Ling}, \bibinfo{person}{Lantao Hu}, \bibinfo{person}{Han Li}, {and} \bibinfo{person}{Wayne~Xin Zhao}.} \bibinfo{year}{2025}\natexlab{}.
\newblock \showarticletitle{Generative Recommender with End-to-End Learnable Item Tokenization}. In \bibinfo{booktitle}{\emph{Proceedings of the 48th International {ACM} {SIGIR} Conference on Research and Development in Information Retrieval, {SIGIR} 2025, Padua, Italy, July 13-18, 2025}}, \bibfield{editor}{\bibinfo{person}{Nicola Ferro}, \bibinfo{person}{Maria Maistro}, \bibinfo{person}{Gabriella Pasi}, \bibinfo{person}{Omar Alonso}, \bibinfo{person}{Andrew Trotman}, {and} \bibinfo{person}{Suzan Verberne}} (Eds.). \bibinfo{publisher}{{ACM}}, \bibinfo{pages}{729--739}.
\newblock
\href{https://doi.org/10.1145/3726302.3729989}{doi:\nolinkurl{10.1145/3726302.3729989}}


\bibitem[Liu et~al\mbox{.}(2022)]%
        {TVR_Token_Shift}
\bibfield{author}{\bibinfo{person}{Yuqi Liu}, \bibinfo{person}{Pengfei Xiong}, \bibinfo{person}{Luhui Xu}, \bibinfo{person}{Shengming Cao}, {and} \bibinfo{person}{Qin Jin}.} \bibinfo{year}{2022}\natexlab{}.
\newblock \showarticletitle{TS2-Net: Token Shift and Selection Transformer for Text-Video Retrieval}. In \bibinfo{booktitle}{\emph{Computer Vision - {ECCV} 2022 - 17th European Conference, Tel Aviv, Israel, October 23-27, 2022, Proceedings, Part {XIV}}} \emph{(\bibinfo{series}{Lecture Notes in Computer Science}, Vol.~\bibinfo{volume}{13674})}, \bibfield{editor}{\bibinfo{person}{Shai Avidan}, \bibinfo{person}{Gabriel~J. Brostow}, \bibinfo{person}{Moustapha Ciss{\'{e}}}, \bibinfo{person}{Giovanni~Maria Farinella}, {and} \bibinfo{person}{Tal Hassner}} (Eds.). \bibinfo{publisher}{Springer}, \bibinfo{pages}{319--335}.
\newblock
\href{https://doi.org/10.1007/978-3-031-19781-9\_19}{doi:\nolinkurl{10.1007/978-3-031-19781-9\_19}}


\bibitem[Liu et~al\mbox{.}(2024)]%
        {GR_RecSys3}
\bibfield{author}{\bibinfo{person}{Zihan Liu}, \bibinfo{person}{Yupeng Hou}, {and} \bibinfo{person}{Julian~J. McAuley}.} \bibinfo{year}{2024}\natexlab{}.
\newblock \showarticletitle{Multi-Behavior Generative Recommendation}. In \bibinfo{booktitle}{\emph{Proceedings of the 33rd {ACM} International Conference on Information and Knowledge Management, {CIKM} 2024, Boise, ID, USA, October 21-25, 2024}}, \bibfield{editor}{\bibinfo{person}{Edoardo Serra} {and} \bibinfo{person}{Francesca Spezzano}} (Eds.). \bibinfo{publisher}{{ACM}}, \bibinfo{pages}{1575--1585}.
\newblock
\href{https://doi.org/10.1145/3627673.3679730}{doi:\nolinkurl{10.1145/3627673.3679730}}


\bibitem[Loshchilov and Hutter(2019)]%
        {AdamW}
\bibfield{author}{\bibinfo{person}{Ilya Loshchilov} {and} \bibinfo{person}{Frank Hutter}.} \bibinfo{year}{2019}\natexlab{}.
\newblock \showarticletitle{Decoupled Weight Decay Regularization}. In \bibinfo{booktitle}{\emph{{ICLR} (Poster)}}. \bibinfo{publisher}{OpenReview.net}.
\newblock


\bibitem[Luo et~al\mbox{.}(2022)]%
        {TVR_CLIP4clip}
\bibfield{author}{\bibinfo{person}{Huaishao Luo}, \bibinfo{person}{Lei Ji}, \bibinfo{person}{Ming Zhong}, \bibinfo{person}{Yang Chen}, \bibinfo{person}{Wen Lei}, \bibinfo{person}{Nan Duan}, {and} \bibinfo{person}{Tianrui Li}.} \bibinfo{year}{2022}\natexlab{}.
\newblock \showarticletitle{CLIP4Clip: An empirical study of {CLIP} for end to end video clip retrieval and captioning}.
\newblock \bibinfo{journal}{\emph{Neurocomputing}}  \bibinfo{volume}{508} (\bibinfo{year}{2022}), \bibinfo{pages}{293--304}.
\newblock
\href{https://doi.org/10.1016/J.NEUCOM.2022.07.028}{doi:\nolinkurl{10.1016/J.NEUCOM.2022.07.028}}


\bibitem[Radford et~al\mbox{.}(2021)]%
        {CLIP}
\bibfield{author}{\bibinfo{person}{Alec Radford}, \bibinfo{person}{Jong~Wook Kim}, \bibinfo{person}{Chris Hallacy}, \bibinfo{person}{Aditya Ramesh}, \bibinfo{person}{Gabriel Goh}, \bibinfo{person}{Sandhini Agarwal}, \bibinfo{person}{Girish Sastry}, \bibinfo{person}{Amanda Askell}, \bibinfo{person}{Pamela Mishkin}, \bibinfo{person}{Jack Clark}, \bibinfo{person}{Gretchen Krueger}, {and} \bibinfo{person}{Ilya Sutskever}.} \bibinfo{year}{2021}\natexlab{}.
\newblock \showarticletitle{Learning Transferable Visual Models From Natural Language Supervision}. In \bibinfo{booktitle}{\emph{Proceedings of the 38th International Conference on Machine Learning, {ICML} 2021, 18-24 July 2021, Virtual Event}} \emph{(\bibinfo{series}{Proceedings of Machine Learning Research}, Vol.~\bibinfo{volume}{139})}, \bibfield{editor}{\bibinfo{person}{Marina Meila} {and} \bibinfo{person}{Tong Zhang}} (Eds.). \bibinfo{publisher}{{PMLR}}, \bibinfo{pages}{8748--8763}.
\newblock
\urldef\tempurl%
\url{http://proceedings.mlr.press/v139/radford21a.html}
\showURL{%
\tempurl}


\bibitem[Raffel et~al\mbox{.}(2020)]%
        {T5}
\bibfield{author}{\bibinfo{person}{Colin Raffel}, \bibinfo{person}{Noam Shazeer}, \bibinfo{person}{Adam Roberts}, \bibinfo{person}{Katherine Lee}, \bibinfo{person}{Sharan Narang}, \bibinfo{person}{Michael Matena}, \bibinfo{person}{Yanqi Zhou}, \bibinfo{person}{Wei Li}, {and} \bibinfo{person}{Peter~J. Liu}.} \bibinfo{year}{2020}\natexlab{}.
\newblock \showarticletitle{Exploring the Limits of Transfer Learning with a Unified Text-to-Text Transformer}.
\newblock \bibinfo{journal}{\emph{J. Mach. Learn. Res.}}  \bibinfo{volume}{21} (\bibinfo{year}{2020}), \bibinfo{pages}{140:1--140:67}.
\newblock
\urldef\tempurl%
\url{https://jmlr.org/papers/v21/20-074.html}
\showURL{%
\tempurl}


\bibitem[Rajput et~al\mbox{.}(2023)]%
        {GR_RecSys_TIGER}
\bibfield{author}{\bibinfo{person}{Shashank Rajput}, \bibinfo{person}{Nikhil Mehta}, \bibinfo{person}{Anima Singh}, \bibinfo{person}{Raghunandan~Hulikal Keshavan}, \bibinfo{person}{Trung Vu}, \bibinfo{person}{Lukasz Heldt}, \bibinfo{person}{Lichan Hong}, \bibinfo{person}{Yi Tay}, \bibinfo{person}{Vinh~Q. Tran}, \bibinfo{person}{Jonah Samost}, \bibinfo{person}{Maciej Kula}, \bibinfo{person}{Ed~H. Chi}, {and} \bibinfo{person}{Mahesh Sathiamoorthy}.} \bibinfo{year}{2023}\natexlab{}.
\newblock \showarticletitle{Recommender Systems with Generative Retrieval}. In \bibinfo{booktitle}{\emph{Advances in Neural Information Processing Systems 36: Annual Conference on Neural Information Processing Systems 2023, NeurIPS 2023, New Orleans, LA, USA, December 10 - 16, 2023}}, \bibfield{editor}{\bibinfo{person}{Alice Oh}, \bibinfo{person}{Tristan Naumann}, \bibinfo{person}{Amir Globerson}, \bibinfo{person}{Kate Saenko}, \bibinfo{person}{Moritz Hardt}, {and} \bibinfo{person}{Sergey Levine}} (Eds.).
\newblock
\urldef\tempurl%
\url{http://papers.nips.cc/paper\_files/paper/2023/hash/20dcab0f14046a5c6b02b61da9f13229-Abstract-Conference.html}
\showURL{%
\tempurl}


\bibitem[Robertson and Zaragoza(2009)]%
        {BM25}
\bibfield{author}{\bibinfo{person}{Stephen~E. Robertson} {and} \bibinfo{person}{Hugo Zaragoza}.} \bibinfo{year}{2009}\natexlab{}.
\newblock \showarticletitle{The Probabilistic Relevance Framework: {BM25} and Beyond}.
\newblock \bibinfo{journal}{\emph{Found. Trends Inf. Retr.}} \bibinfo{volume}{3}, \bibinfo{number}{4} (\bibinfo{year}{2009}), \bibinfo{pages}{333--389}.
\newblock


\bibitem[Rohrbach et~al\mbox{.}(2017)]%
        {LSMDC}
\bibfield{author}{\bibinfo{person}{Anna Rohrbach}, \bibinfo{person}{Atousa Torabi}, \bibinfo{person}{Marcus Rohrbach}, \bibinfo{person}{Niket Tandon}, \bibinfo{person}{Christopher~Joseph Pal}, \bibinfo{person}{Hugo Larochelle}, \bibinfo{person}{Aaron~C. Courville}, {and} \bibinfo{person}{Bernt Schiele}.} \bibinfo{year}{2017}\natexlab{}.
\newblock \showarticletitle{Movie Description}.
\newblock \bibinfo{journal}{\emph{Int. J. Comput. Vis.}} \bibinfo{volume}{123}, \bibinfo{number}{1} (\bibinfo{year}{2017}), \bibinfo{pages}{94--120}.
\newblock
\href{https://doi.org/10.1007/S11263-016-0987-1}{doi:\nolinkurl{10.1007/S11263-016-0987-1}}


\bibitem[Si et~al\mbox{.}(2023)]%
        {GR_RecSys1}
\bibfield{author}{\bibinfo{person}{Zihua Si}, \bibinfo{person}{Zhongxiang Sun}, \bibinfo{person}{Jiale Chen}, \bibinfo{person}{Guozhang Chen}, \bibinfo{person}{Xiaoxue Zang}, \bibinfo{person}{Kai Zheng}, \bibinfo{person}{Yang Song}, \bibinfo{person}{Xiao Zhang}, {and} \bibinfo{person}{Jun Xu}.} \bibinfo{year}{2023}\natexlab{}.
\newblock \showarticletitle{Generative Retrieval with Semantic Tree-Structured Item Identifiers via Contrastive Learning}.
\newblock \bibinfo{journal}{\emph{CoRR}}  \bibinfo{volume}{abs/2309.13375} (\bibinfo{year}{2023}).
\newblock
\showeprint[arXiv]{2309.13375}
\href{https://doi.org/10.48550/ARXIV.2309.13375}{doi:\nolinkurl{10.48550/ARXIV.2309.13375}}


\bibitem[Sun et~al\mbox{.}(2023)]%
        {GR_DR_1}
\bibfield{author}{\bibinfo{person}{Weiwei Sun}, \bibinfo{person}{Lingyong Yan}, \bibinfo{person}{Zheng Chen}, \bibinfo{person}{Shuaiqiang Wang}, \bibinfo{person}{Haichao Zhu}, \bibinfo{person}{Pengjie Ren}, \bibinfo{person}{Zhumin Chen}, \bibinfo{person}{Dawei Yin}, \bibinfo{person}{Maarten de Rijke}, {and} \bibinfo{person}{Zhaochun Ren}.} \bibinfo{year}{2023}\natexlab{}.
\newblock \showarticletitle{Learning to Tokenize for Generative Retrieval}. In \bibinfo{booktitle}{\emph{Advances in Neural Information Processing Systems 36: Annual Conference on Neural Information Processing Systems 2023, NeurIPS 2023, New Orleans, LA, USA, December 10 - 16, 2023}}, \bibfield{editor}{\bibinfo{person}{Alice Oh}, \bibinfo{person}{Tristan Naumann}, \bibinfo{person}{Amir Globerson}, \bibinfo{person}{Kate Saenko}, \bibinfo{person}{Moritz Hardt}, {and} \bibinfo{person}{Sergey Levine}} (Eds.).
\newblock
\urldef\tempurl%
\url{http://papers.nips.cc/paper\_files/paper/2023/hash/91228b942a4528cdae031c1b68b127e8-Abstract-Conference.html}
\showURL{%
\tempurl}


\bibitem[Tay et~al\mbox{.}(2022)]%
        {GR_DR_DSI}
\bibfield{author}{\bibinfo{person}{Yi Tay}, \bibinfo{person}{Vinh Tran}, \bibinfo{person}{Mostafa Dehghani}, \bibinfo{person}{Jianmo Ni}, \bibinfo{person}{Dara Bahri}, \bibinfo{person}{Harsh Mehta}, \bibinfo{person}{Zhen Qin}, \bibinfo{person}{Kai Hui}, \bibinfo{person}{Zhe Zhao}, \bibinfo{person}{Jai~Prakash Gupta}, \bibinfo{person}{Tal Schuster}, \bibinfo{person}{William~W. Cohen}, {and} \bibinfo{person}{Donald Metzler}.} \bibinfo{year}{2022}\natexlab{}.
\newblock \showarticletitle{Transformer Memory as a Differentiable Search Index}. In \bibinfo{booktitle}{\emph{Advances in Neural Information Processing Systems 35: Annual Conference on Neural Information Processing Systems 2022, NeurIPS 2022, New Orleans, LA, USA, November 28 - December 9, 2022}}, \bibfield{editor}{\bibinfo{person}{Sanmi Koyejo}, \bibinfo{person}{S.~Mohamed}, \bibinfo{person}{A.~Agarwal}, \bibinfo{person}{Danielle Belgrave}, \bibinfo{person}{K.~Cho}, {and} \bibinfo{person}{A.~Oh}} (Eds.).
\newblock
\urldef\tempurl%
\url{http://papers.nips.cc/paper\_files/paper/2022/hash/892840a6123b5ec99ebaab8be1530fba-Abstract-Conference.html}
\showURL{%
\tempurl}


\bibitem[Tian et~al\mbox{.}(2024)]%
        {TVR_EERCF}
\bibfield{author}{\bibinfo{person}{Kaibin Tian}, \bibinfo{person}{Yanhua Cheng}, \bibinfo{person}{Yi Liu}, \bibinfo{person}{Xinglin Hou}, \bibinfo{person}{Quan Chen}, {and} \bibinfo{person}{Han Li}.} \bibinfo{year}{2024}\natexlab{}.
\newblock \showarticletitle{Towards Efficient and Effective Text-to-Video Retrieval with Coarse-to-Fine Visual Representation Learning}. In \bibinfo{booktitle}{\emph{Thirty-Eighth {AAAI} Conference on Artificial Intelligence, {AAAI} 2024, Thirty-Sixth Conference on Innovative Applications of Artificial Intelligence, {IAAI} 2024, Fourteenth Symposium on Educational Advances in Artificial Intelligence, {EAAI} 2014, February 20-27, 2024, Vancouver, Canada}}, \bibfield{editor}{\bibinfo{person}{Michael~J. Wooldridge}, \bibinfo{person}{Jennifer~G. Dy}, {and} \bibinfo{person}{Sriraam Natarajan}} (Eds.). \bibinfo{publisher}{{AAAI} Press}, \bibinfo{pages}{5207--5214}.
\newblock
\href{https://doi.org/10.1609/AAAI.V38I6.28327}{doi:\nolinkurl{10.1609/AAAI.V38I6.28327}}


\bibitem[Vaswani et~al\mbox{.}(2017)]%
        {Transformer}
\bibfield{author}{\bibinfo{person}{Ashish Vaswani}, \bibinfo{person}{Noam Shazeer}, \bibinfo{person}{Niki Parmar}, \bibinfo{person}{Jakob Uszkoreit}, \bibinfo{person}{Llion Jones}, \bibinfo{person}{Aidan~N. Gomez}, \bibinfo{person}{Lukasz Kaiser}, {and} \bibinfo{person}{Illia Polosukhin}.} \bibinfo{year}{2017}\natexlab{}.
\newblock \showarticletitle{Attention is All you Need}. In \bibinfo{booktitle}{\emph{Advances in Neural Information Processing Systems 30: Annual Conference on Neural Information Processing Systems 2017, December 4-9, 2017, Long Beach, CA, {USA}}}, \bibfield{editor}{\bibinfo{person}{Isabelle Guyon}, \bibinfo{person}{Ulrike von Luxburg}, \bibinfo{person}{Samy Bengio}, \bibinfo{person}{Hanna~M. Wallach}, \bibinfo{person}{Rob Fergus}, \bibinfo{person}{S.~V.~N. Vishwanathan}, {and} \bibinfo{person}{Roman Garnett}} (Eds.). \bibinfo{pages}{5998--6008}.
\newblock
\urldef\tempurl%
\url{https://proceedings.neurips.cc/paper/2017/hash/3f5ee243547dee91fbd053c1c4a845aa-Abstract.html}
\showURL{%
\tempurl}


\bibitem[Wang et~al\mbox{.}(2025a)]%
        {CV}
\bibfield{author}{\bibinfo{person}{Tianqi Wang}, \bibinfo{person}{Jingcai Guo}, \bibinfo{person}{Depeng Li}, {and} \bibinfo{person}{Zhi Chen}.} \bibinfo{year}{2025}\natexlab{a}.
\newblock \showarticletitle{On the Discrimination and Consistency for Exemplar-Free Class Incremental Learning}. In \bibinfo{booktitle}{\emph{{IJCAI}}}. \bibinfo{publisher}{ijcai.org}, \bibinfo{pages}{6424--6432}.
\newblock


\bibitem[Wang et~al\mbox{.}(2022a)]%
        {GR_DR_NCI}
\bibfield{author}{\bibinfo{person}{Yujing Wang}, \bibinfo{person}{Yingyan Hou}, \bibinfo{person}{Haonan Wang}, \bibinfo{person}{Ziming Miao}, \bibinfo{person}{Shibin Wu}, \bibinfo{person}{Qi Chen}, \bibinfo{person}{Yuqing Xia}, \bibinfo{person}{Chengmin Chi}, \bibinfo{person}{Guoshuai Zhao}, \bibinfo{person}{Zheng Liu}, \bibinfo{person}{Xing Xie}, \bibinfo{person}{Hao Sun}, \bibinfo{person}{Weiwei Deng}, \bibinfo{person}{Qi Zhang}, {and} \bibinfo{person}{Mao Yang}.} \bibinfo{year}{2022}\natexlab{a}.
\newblock \showarticletitle{A Neural Corpus Indexer for Document Retrieval}. In \bibinfo{booktitle}{\emph{Advances in Neural Information Processing Systems 35: Annual Conference on Neural Information Processing Systems 2022, NeurIPS 2022, New Orleans, LA, USA, November 28 - December 9, 2022}}, \bibfield{editor}{\bibinfo{person}{Sanmi Koyejo}, \bibinfo{person}{S.~Mohamed}, \bibinfo{person}{A.~Agarwal}, \bibinfo{person}{Danielle Belgrave}, \bibinfo{person}{K.~Cho}, {and} \bibinfo{person}{A.~Oh}} (Eds.).
\newblock
\urldef\tempurl%
\url{http://papers.nips.cc/paper\_files/paper/2022/hash/a46156bd3579c3b268108ea6aca71d13-Abstract-Conference.html}
\showURL{%
\tempurl}


\bibitem[Wang et~al\mbox{.}(2024)]%
        {InternVideo2}
\bibfield{author}{\bibinfo{person}{Yi Wang}, \bibinfo{person}{Kunchang Li}, \bibinfo{person}{Xinhao Li}, \bibinfo{person}{Jiashuo Yu}, \bibinfo{person}{Yinan He}, \bibinfo{person}{Guo Chen}, \bibinfo{person}{Baoqi Pei}, \bibinfo{person}{Rongkun Zheng}, \bibinfo{person}{Zun Wang}, \bibinfo{person}{Yansong Shi}, \bibinfo{person}{Tianxiang Jiang}, \bibinfo{person}{Songze Li}, \bibinfo{person}{Jilan Xu}, \bibinfo{person}{Hongjie Zhang}, \bibinfo{person}{Yifei Huang}, \bibinfo{person}{Yu Qiao}, \bibinfo{person}{Yali Wang}, {and} \bibinfo{person}{Limin Wang}.} \bibinfo{year}{2024}\natexlab{}.
\newblock \showarticletitle{InternVideo2: Scaling Foundation Models for Multimodal Video Understanding}. In \bibinfo{booktitle}{\emph{Computer Vision - {ECCV} 2024 - 18th European Conference, Milan, Italy, September 29-October 4, 2024, Proceedings, Part {LXXXV}}} \emph{(\bibinfo{series}{Lecture Notes in Computer Science}, Vol.~\bibinfo{volume}{15143})}, \bibfield{editor}{\bibinfo{person}{Ales Leonardis}, \bibinfo{person}{Elisa Ricci}, \bibinfo{person}{Stefan Roth}, \bibinfo{person}{Olga Russakovsky}, \bibinfo{person}{Torsten Sattler}, {and} \bibinfo{person}{G{\"{u}}l Varol}} (Eds.). \bibinfo{publisher}{Springer}, \bibinfo{pages}{396--416}.
\newblock
\href{https://doi.org/10.1007/978-3-031-73013-9\_23}{doi:\nolinkurl{10.1007/978-3-031-73013-9\_23}}


\bibitem[Wang et~al\mbox{.}(2022b)]%
        {InternVideo}
\bibfield{author}{\bibinfo{person}{Yi Wang}, \bibinfo{person}{Kunchang Li}, \bibinfo{person}{Yizhuo Li}, \bibinfo{person}{Yinan He}, \bibinfo{person}{Bingkun Huang}, \bibinfo{person}{Zhiyu Zhao}, \bibinfo{person}{Hongjie Zhang}, \bibinfo{person}{Jilan Xu}, \bibinfo{person}{Yi Liu}, \bibinfo{person}{Zun Wang}, \bibinfo{person}{Sen Xing}, \bibinfo{person}{Guo Chen}, \bibinfo{person}{Junting Pan}, \bibinfo{person}{Jiashuo Yu}, \bibinfo{person}{Yali Wang}, \bibinfo{person}{Limin Wang}, {and} \bibinfo{person}{Yu Qiao}.} \bibinfo{year}{2022}\natexlab{b}.
\newblock \showarticletitle{InternVideo: General Video Foundation Models via Generative and Discriminative Learning}.
\newblock \bibinfo{journal}{\emph{CoRR}}  \bibinfo{volume}{abs/2212.03191} (\bibinfo{year}{2022}).
\newblock
\showeprint[arXiv]{2212.03191}
\href{https://doi.org/10.48550/ARXIV.2212.03191}{doi:\nolinkurl{10.48550/ARXIV.2212.03191}}


\bibitem[Wang et~al\mbox{.}(2025b)]%
        {InternVideo2.5}
\bibfield{author}{\bibinfo{person}{Yi Wang}, \bibinfo{person}{Xinhao Li}, \bibinfo{person}{Ziang Yan}, \bibinfo{person}{Yinan He}, \bibinfo{person}{Jiashuo Yu}, \bibinfo{person}{Xiangyu Zeng}, \bibinfo{person}{Chenting Wang}, \bibinfo{person}{Changlian Ma}, \bibinfo{person}{Haian Huang}, \bibinfo{person}{Jianfei Gao}, \bibinfo{person}{Min Dou}, \bibinfo{person}{Kai Chen}, \bibinfo{person}{Wenhai Wang}, \bibinfo{person}{Yu Qiao}, \bibinfo{person}{Yali Wang}, {and} \bibinfo{person}{Limin Wang}.} \bibinfo{year}{2025}\natexlab{b}.
\newblock \showarticletitle{InternVideo2.5: Empowering Video MLLMs with Long and Rich Context Modeling}.
\newblock \bibinfo{journal}{\emph{CoRR}}  \bibinfo{volume}{abs/2501.12386} (\bibinfo{year}{2025}).
\newblock
\showeprint[arXiv]{2501.12386}
\href{https://doi.org/10.48550/ARXIV.2501.12386}{doi:\nolinkurl{10.48550/ARXIV.2501.12386}}


\bibitem[Xu et~al\mbox{.}(2016)]%
        {MSRVTT}
\bibfield{author}{\bibinfo{person}{Jun Xu}, \bibinfo{person}{Tao Mei}, \bibinfo{person}{Ting Yao}, {and} \bibinfo{person}{Yong Rui}.} \bibinfo{year}{2016}\natexlab{}.
\newblock \showarticletitle{{MSR-VTT:} {A} Large Video Description Dataset for Bridging Video and Language}. In \bibinfo{booktitle}{\emph{2016 {IEEE} Conference on Computer Vision and Pattern Recognition, {CVPR} 2016, Las Vegas, NV, USA, June 27-30, 2016}}. \bibinfo{publisher}{{IEEE} Computer Society}, \bibinfo{pages}{5288--5296}.
\newblock
\href{https://doi.org/10.1109/CVPR.2016.571}{doi:\nolinkurl{10.1109/CVPR.2016.571}}


\bibitem[Xue et~al\mbox{.}(2022)]%
        {TVR_CLIPViP}
\bibfield{author}{\bibinfo{person}{Hongwei Xue}, \bibinfo{person}{Yuchong Sun}, \bibinfo{person}{Bei Liu}, \bibinfo{person}{Jianlong Fu}, \bibinfo{person}{Ruihua Song}, \bibinfo{person}{Houqiang Li}, {and} \bibinfo{person}{Jiebo Luo}.} \bibinfo{year}{2022}\natexlab{}.
\newblock \showarticletitle{CLIP-ViP: Adapting Pre-trained Image-Text Model to Video-Language Representation Alignment}.
\newblock \bibinfo{journal}{\emph{CoRR}}  \bibinfo{volume}{abs/2209.06430} (\bibinfo{year}{2022}).
\newblock
\showeprint[arXiv]{2209.06430}
\href{https://doi.org/10.48550/ARXIV.2209.06430}{doi:\nolinkurl{10.48550/ARXIV.2209.06430}}


\bibitem[Xue et~al\mbox{.}(2024)]%
        {VLM4}
\bibfield{author}{\bibinfo{person}{Le Xue}, \bibinfo{person}{Manli Shu}, \bibinfo{person}{Anas Awadalla}, \bibinfo{person}{Jun Wang}, \bibinfo{person}{An Yan}, \bibinfo{person}{Senthil Purushwalkam}, \bibinfo{person}{Honglu Zhou}, \bibinfo{person}{Viraj Prabhu}, \bibinfo{person}{Yutong Dai}, \bibinfo{person}{Michael~S. Ryoo}, \bibinfo{person}{Shrikant Kendre}, \bibinfo{person}{Jieyu Zhang}, \bibinfo{person}{Can Qin}, \bibinfo{person}{Shu Zhang}, \bibinfo{person}{Chia{-}Chih Chen}, \bibinfo{person}{Ning Yu}, \bibinfo{person}{Juntao Tan}, \bibinfo{person}{Tulika~Manoj Awalgaonkar}, \bibinfo{person}{Shelby Heinecke}, \bibinfo{person}{Huan Wang}, \bibinfo{person}{Yejin Choi}, \bibinfo{person}{Ludwig Schmidt}, \bibinfo{person}{Zeyuan Chen}, \bibinfo{person}{Silvio Savarese}, \bibinfo{person}{Juan~Carlos Niebles}, \bibinfo{person}{Caiming Xiong}, {and} \bibinfo{person}{Ran Xu}.} \bibinfo{year}{2024}\natexlab{}.
\newblock \showarticletitle{xGen-MM {(BLIP-3):} {A} Family of Open Large Multimodal Models}.
\newblock \bibinfo{journal}{\emph{CoRR}}  \bibinfo{volume}{abs/2408.08872} (\bibinfo{year}{2024}).
\newblock
\showeprint[arXiv]{2408.08872}
\href{https://doi.org/10.48550/ARXIV.2408.08872}{doi:\nolinkurl{10.48550/ARXIV.2408.08872}}


\bibitem[Zhai et~al\mbox{.}(2025)]%
        {GR_MM1}
\bibfield{author}{\bibinfo{person}{Jianyang Zhai}, \bibinfo{person}{Zi{-}Feng Mai}, \bibinfo{person}{Chang{-}Dong Wang}, \bibinfo{person}{Feidiao Yang}, \bibinfo{person}{Xiawu Zheng}, \bibinfo{person}{Hui Li}, {and} \bibinfo{person}{Yonghong Tian}.} \bibinfo{year}{2025}\natexlab{}.
\newblock \showarticletitle{Multimodal Quantitative Language for Generative Recommendation}. In \bibinfo{booktitle}{\emph{The Thirteenth International Conference on Learning Representations, {ICLR} 2025, Singapore, April 24-28, 2025}}. \bibinfo{publisher}{OpenReview.net}.
\newblock
\urldef\tempurl%
\url{https://openreview.net/forum?id=v7YrIjpkTF}
\showURL{%
\tempurl}


\bibitem[Zhang et~al\mbox{.}(2024)]%
        {GR_IR_IRGEN}
\bibfield{author}{\bibinfo{person}{Yidan Zhang}, \bibinfo{person}{Ting Zhang}, \bibinfo{person}{Dong Chen}, \bibinfo{person}{Yujing Wang}, \bibinfo{person}{Qi Chen}, \bibinfo{person}{Xing Xie}, \bibinfo{person}{Hao Sun}, \bibinfo{person}{Weiwei Deng}, \bibinfo{person}{Qi Zhang}, \bibinfo{person}{Fan Yang}, \bibinfo{person}{Mao Yang}, \bibinfo{person}{Qingmin Liao}, \bibinfo{person}{Jingdong Wang}, {and} \bibinfo{person}{Baining Guo}.} \bibinfo{year}{2024}\natexlab{}.
\newblock \showarticletitle{IRGen: Generative Modeling for Image Retrieval}. In \bibinfo{booktitle}{\emph{Computer Vision - {ECCV} 2024 - 18th European Conference, Milan, Italy, September 29-October 4, 2024, Proceedings, Part {XV}}} \emph{(\bibinfo{series}{Lecture Notes in Computer Science}, Vol.~\bibinfo{volume}{15073})}, \bibfield{editor}{\bibinfo{person}{Ales Leonardis}, \bibinfo{person}{Elisa Ricci}, \bibinfo{person}{Stefan Roth}, \bibinfo{person}{Olga Russakovsky}, \bibinfo{person}{Torsten Sattler}, {and} \bibinfo{person}{G{\"{u}}l Varol}} (Eds.). \bibinfo{publisher}{Springer}, \bibinfo{pages}{21--41}.
\newblock
\href{https://doi.org/10.1007/978-3-031-72633-0\_2}{doi:\nolinkurl{10.1007/978-3-031-72633-0\_2}}


\bibitem[Zhang et~al\mbox{.}(2025)]%
        {CLIP2}
\bibfield{author}{\bibinfo{person}{Zhizhen Zhang}, \bibinfo{person}{Lei Zhu}, \bibinfo{person}{Zhen Fang}, \bibinfo{person}{Zi Huang}, {and} \bibinfo{person}{Yadan Luo}.} \bibinfo{year}{2025}\natexlab{}.
\newblock \showarticletitle{Provable Ordering and Continuity in Vision-Language Pretraining for Generalizable Embodied Agents}.
\newblock \bibinfo{journal}{\emph{CoRR}}  \bibinfo{volume}{abs/2502.01218} (\bibinfo{year}{2025}).
\newblock


\bibitem[Zhao et~al\mbox{.}(2025a)]%
        {TVR_CTVR}
\bibfield{author}{\bibinfo{person}{Zecheng Zhao}, \bibinfo{person}{Zhi Chen}, \bibinfo{person}{Zi Huang}, \bibinfo{person}{Shazia Sadiq}, {and} \bibinfo{person}{Tong Chen}.} \bibinfo{year}{2025}\natexlab{a}.
\newblock \showarticletitle{Continual Text-to-Video Retrieval with Frame Fusion and Task-Aware Routing}. In \bibinfo{booktitle}{\emph{Proceedings of the 48th International {ACM} {SIGIR} Conference on Research and Development in Information Retrieval, {SIGIR} 2025, Padua, Italy, July 13-18, 2025}}, \bibfield{editor}{\bibinfo{person}{Nicola Ferro}, \bibinfo{person}{Maria Maistro}, \bibinfo{person}{Gabriella Pasi}, \bibinfo{person}{Omar Alonso}, \bibinfo{person}{Andrew Trotman}, {and} \bibinfo{person}{Suzan Verberne}} (Eds.). \bibinfo{publisher}{{ACM}}, \bibinfo{pages}{1011--1021}.
\newblock
\href{https://doi.org/10.1145/3726302.3729936}{doi:\nolinkurl{10.1145/3726302.3729936}}


\bibitem[Zhao et~al\mbox{.}(2025b)]%
        {TVR_SynTVR}
\bibfield{author}{\bibinfo{person}{Zecheng Zhao}, \bibinfo{person}{Selena Song}, \bibinfo{person}{Tong Chen}, \bibinfo{person}{Zhi Chen}, \bibinfo{person}{Shazia Sadiq}, {and} \bibinfo{person}{Yadan Luo}.} \bibinfo{year}{2025}\natexlab{b}.
\newblock \showarticletitle{Are Synthetic Videos Useful? {A} Benchmark for Retrieval-Centric Evaluation of Synthetic Videos}.
\newblock \bibinfo{journal}{\emph{CoRR}}  \bibinfo{volume}{abs/2507.02316} (\bibinfo{year}{2025}).
\newblock
\showeprint[arXiv]{2507.02316}
\href{https://doi.org/10.48550/ARXIV.2507.02316}{doi:\nolinkurl{10.48550/ARXIV.2507.02316}}


\end{thebibliography}

\end{document}